\RequirePackage{booktabs}
\documentclass[pdflatex,sn-mathphys-num]{sn-jnl}

\usepackage{amsmath,amssymb,amsfonts}%
\usepackage{amsthm}%
\usepackage{mathrsfs}%
\usepackage[title]{appendix}%
\usepackage{textcomp}%
\usepackage{manyfoot}%
\usepackage{algorithm}%
\usepackage{algorithmicx}%
\usepackage{algpseudocode}%
\usepackage{listings}%

\usepackage{array}
\usepackage[caption=false,font=normalsize,labelfont=sf,textfont=sf]{subfig}
\usepackage{stfloats}
\usepackage{url}
\usepackage{verbatim}
\usepackage{adjustbox,lipsum}
\usepackage{pdflscape}
\usepackage[longtable]{multirow}
\usepackage{longtable}
\usepackage[T1]{fontenc}
\usepackage[utf8]{inputenc}
\usepackage{calc}
\usepackage{indentfirst}
\usepackage{fancyhdr}
\usepackage{epstopdf}
\usepackage{lastpage}
\usepackage{ifthen}
\usepackage{float}
\usepackage{amssymb} 
\usepackage[right]{lineno}
\usepackage{setspace}
\usepackage{enumitem}
\usepackage{mathpazo}
\usepackage{titlesec}
\usepackage{etoolbox} 
\usepackage{tabto} 
\usepackage{soul} 

\usepackage{multirow}
\usepackage{microtype} 
\usepackage{tikz} 
\usepackage{totcount} 
\usepackage{changepage} 
\usepackage{attrib} 
\usepackage{tabularx}
\usepackage{pbox} 
\usepackage{ragged2e} 
\usepackage[subfigure]{tocloft} 
\usepackage{marginfix} 
\usepackage{enotez} 
\usepackage{mathtools}
\usepackage{forest}
\usepackage{environ}         
\usepackage{svg}

\newlength{\myl}
\let\origequation=\equation
\let\origendequation=\endequation
\RenewEnviron{equation}{
  \settowidth{\myl}{$\BODY$}                       
  \origequation
  \ifdimcomp{\the\linewidth}{>}{\the\myl}
  {\ensuremath{\BODY}}                             
  {\resizebox{\linewidth}{!}{\ensuremath{\BODY}}}  
  \origendequation
}

\usetikzlibrary{trees,positioning,shapes,shadows,arrows.meta}

\theoremstyle{thmstyleone}%
\theoremstyle{thmstyletwo}%

\theoremstyle{thmstylethree}%

\begin{document}

\title[Article Title]{Beyond Traditional Single Object Tracking: A Survey}


\author*[1]{\fnm{Omar} \sur{Abdelaziz}}\email{oabdelaz@student.ubc.ca}

\author[1]{\fnm{Mohamed} \sur{Shehata}}\email{mohamed.sami.shehata@ubc.ca}

\author[1]{\fnm{Mohamed} \sur{Mohamed}}\email{mohamed.abdelpakey@ubc.ca}

\affil[1]{\orgdiv{Department of Computer Science, Mathematics, Physics and Statistics}, \orgname{The University of British Columbia}, \orgaddress{\street{3333 University Way}, \city{Kelowna}, \postcode{V1V1V7}, \state{BC}, \country{Canada}}}


\abstract{Single object tracking is a vital task of many applications in critical fields. However, it is still considered one of the most challenging vision tasks. In recent years, computer vision, especially object tracking, witnessed the introduction or adoption of many novel techniques, setting new fronts for performance. In this survey, we visit some of the cutting-edge techniques in vision, such as Sequence Models, Generative Models, Self-supervised Learning, Unsupervised Learning, Reinforcement Learning, Meta-Learning, Continual Learning, and Domain Adaptation, focusing on their application in single object tracking. We propose a novel categorization of single object tracking methods based on novel techniques and trends. Also, we conduct a comparative analysis of the performance reported by the methods presented on popular tracking benchmarks. Moreover, we analyze the pros and cons of the presented approaches and present a guide for non-traditional techniques in single object tracking. Finally, we suggest potential avenues for future research in single-object tracking.}




\maketitle

\section{Introduction}

Single object tracking has been a fundamental task in computer vision for a long time. The single object tracking model is designed to process a sequence of images as its input. The first frame of the sequence is annotated with a bounding box around the object of interest, and the model finds the trajectory of the locations throughout the sequence.

From an application standpoint, single object tracking is serving as a foundational cornerstone in computer vision that represents the linchpin of video surveillance \cite{Mathur2018Intelligent}, autonomous driving \cite{cao2021robust}, video analysis \cite{zheng2023cows} and scene understanding \cite{chen2020mvlidarnet}. In the dynamic landscape of single object tracking, the ability to follow and monitor a specific entity across a sequence of frames is paramount. Yet, the intrinsic challenges embedded in real-world scenarios, including variations in scale, pose, illumination, and occlusions, mandate a robust tracking system that can adapt seamlessly to an ever-changing environment. Traditional methods \cite{Yilmaz2006Object, Smeulders2014Visual}, reliant on handcrafted features and heuristic assumptions, often fail to address the previous challenges and result in low accuracy, adaptability, and efficiency posed by modern tracking applications.

Prior reviews on single object tracking primarily focused on different methods without highlighting new paradigms and methodologies that emerged and their effect on single object tracking. For instance, \cite{Yilmaz2006Object} provides a good base for categorizing the different methods of the time of the work as well as providing clear definitions for challenges inherent in single object tracking. Another work \cite{Smeulders2014Visual} provides extensive experiments on different trackers to assess their performance analytically.
More recently, reviews have been specialized to categorize trackers of the same main paradigm. Specifically, \cite{Javed2023Visual} surveys Siamese networks and discriminative correlation filters. Moreover, with the advent of Transformers revolutionizing most computer vision fields, \cite{kugarajeevan2023transformers} categorizes the tracking methodologies according to the employment of the Transformers and provides experiments with insights into the performance of different methods across various datasets.

\begin{figure}    
    \scalebox{0.6}{
        \includegraphics[width=1\columnwidth]{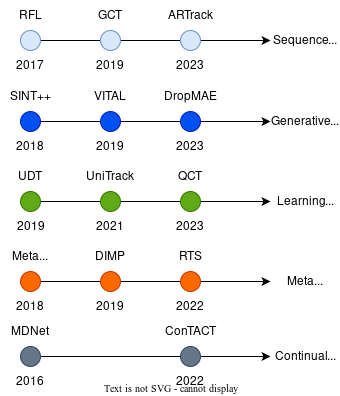}
    }
    \caption{A timeline of the most important trackers per each of the six broad categories of this survey.}
    \label{fig:time}
\end{figure}

This survey paper embarks on a comprehensive exploration of recent advancements in single object tracking, centring on the integration of non-traditional techniques. The landscape of single-object tracking is undergoing a paradigm shift as researchers increasingly leverage advanced machine-learning models and innovative methodologies to overcome the limitations of conventional tracking methods. The focus of this survey is to present an exhaustive review of these non-traditional techniques, shedding light on their underlying principles, strengths, and limitations. We extensively survey the single object tracking methods that utilize cutting-edge deep learning techniques, including Sequence Models, Generative Models, Self-supervised Learning, Unsupervised Learning, Reinforcement Learning, Meta-Learning, Continual Learning, and Domain Adaptation. This survey aims to serve as a comprehensive resource for researchers navigating the complex terrain of single-object tracking. A timeline for the most important methods per each broad category is presented in figure \ref{fig:time}.

This survey poses itself at the intersection of these cutting-edge techniques, extensively examining the state-of-the-art in single object tracking. In the subsequent sections, each non-traditional technique will be dissected, offering an in-depth analysis of methodologies, challenges, and potential directions for future research. Throughout this exploration, readers will gain a comprehensive understanding of the nuanced landscape of single object tracking, equipped with insights to make informed decisions in developing and applying tracking models.

\section{Related Works}
In the existing literature, several survey studies on Visual Object Tracking (VOT) have been conducted over the past two decades. For example, Yilmaz et al. \cite{Yilmaz2006Object} stand out as pioneers in comprehensively analyzing the tracking pipeline. They systematically categorized tracking methods into feature correspondence, primitive geometric models, and contour methods.

Smeulders et al. \cite{Smeulders2014Visual} provided a new dataset, the ALOV++ and conducted an experimental survey that covered $19$ distinct online trackers, spanning the period from $1999$ to $2012$. Their study included multiple evaluation metrics applied to single object trackers. Zhang et al. \cite{ZHANG2021Recent} contributed a survey focused on recent methodologies in deep learning and correlation-filter paradigms. Javed et al. \cite{Javed2023Visual} focused on recent developments in the two most popular paradigms of traditional and deep learning-based methods: discriminative correlation filters and the Siamese-based paradigms. In their survey, $90$ trackers were presented, and $9$ trackers out of them were experimentally tested. 

Given the current state of Vision Transformers (ViTs) \cite{dosovitskiy2021an} and its ubiquitousness in computer vision, Kugarajeevan et al. \cite{kugarajeevan2023transformers} extensively surveyed the new paradigms emerging with the adoption of ViT in single object tracking. They conducted attribute-based comparisons of different transformer-based and non-transformer-based trackers. Each tracking challenge, e.g., camera motion and shape deformation, is examined independently in these comparisons. Moreover, all transformer-based and non-transformer-based methods are compared in another set of experiments to demonstrate the effectiveness of the transformer-based paradigms. 

Contrary to other works, this work sheds light on the emerging theoretical deep learning paradigms and their application in single-object tracking. The main focus of this survey is to delve into recent single-object tracking methods that are built upon a solid foundation of statistical theories and have led to the introduction of new paradigms such as generative models and meta-learning. 

\section{Taxonomy}
\label{sec:Taxonomy}
\begin{figure}
\tikzset{
    basic/.style  = {draw, text width=3cm, align=center, font=\sffamily, rectangle},
    root/.style   = {basic, rounded corners=2pt, thin, align=center, fill=green!30},
    onode/.style = {basic, thin, rounded corners=2pt, align=center, fill=green!60,text width=3cm,},
    tnode/.style = {basic, thin, align=left, fill=pink!60, text width=15em, align=center},
    xnode/.style = {basic, thin, rounded corners=2pt, align=center, fill=blue!20,text width=5cm,},
    wnode/.style = {basic, thin, align=left, fill=pink!10!blue!80!red!10, text width=6.5em},
    edge from parent/.style={draw=black, edge from parent fork right}
}
\scalebox{.45}{
\begin{forest} for tree={
    grow=east,
    growth parent anchor=west,
    parent anchor=east,
    child anchor=west,
    edge path={\noexpand\path[\forestoption{edge},->, >={latex}] 
         (!u.parent anchor) -- +(10pt,0pt) |-  (.child anchor) 
         \forestoption{edge label};}
}
[Non-traditional Tracking Methods, basic,  l sep=10mm,
    [Sequence Models, xnode,  l sep=10mm,
        [Autoregressive Models, tnode, l sep=10mm,
            [{ARTrack \cite{wei2023autoregressive}}, wnode]
            [SeqTrack \cite{Chen2023SeqTrack}, wnode]
            ]
        [Memory-Based Models, tnode, l sep=10mm,
            [AD-LSTM \cite{zhang2023attention}, wnode]
            [Siamese R-RPN \cite{zhao2021cooperative}, wnode]
            [GCT \cite{gao2019graph}, wnode]
            [BA + TID + DDM \cite{wang2023robust}, wnode]
            [RFL \cite{yang2017recurrent}, wnode]]]
    [Generative Probabilistic Models, xnode,  l sep=10mm,
        [Masked Image Modelling, tnode, l sep=10mm,
            [MAT \cite{Zhao2023Representation}, wnode]
            [DropMAE \cite{Wu2023DropMAE}, wnode]]
        [Generative Adversarial Models, tnode, l sep=10mm,
            [TGGAN \cite{Guo2018generating}, wnode],   
            [VITAL \cite{Song2018VITAL}, wnode]
            [GARAT \cite{Yao2022GARAT}, wnode]
            [AFSL \cite{Yin2020Adversarial}, wnode]]
        [Diffusion Models, tnode, l sep=10mm,
            [Diff-SiamRPN++ \cite{Zhang2023Siamese}, wnode]]
        [Variational Autoencoders, tnode, l sep=10mm,
            [SINT++ \cite{wang2018sint++}, wnode]
            [VAE-MCMC \cite{Kwon2020Robust}, wnode]
            ]
            ]
    [Learning Paradigms, xnode,  l sep=10mm,
        [Self-Supervised, tnode, l sep=10mm,
            [QCT \cite{Zhu2023Consistency-based}, wnode]
            [Joint-task \cite{Li2019Joint-task}, wnode]
            [ETC \cite{Zhu2022Exploiting}, wnode]
            [Crop-Transform-Paste \cite{Li2022Self-Supervised}, wnode]
            [CycleSiam \cite{Yuan2020Self-supervised}, wnode]
            [UniTrack \cite{Wang2021Do}, wnode]]
        [Unsupervised, tnode, l sep=10mm,
            [PUL \cite{Wu2021Progressive}, wnode]
            [UDT \cite{Wang2019Unsupervised}, wnode]
            ]
        ]
    [Meta-Learning, xnode,  l sep=10mm,
        [Meta-Tracker \cite{Park2018Meta-Tracker}, wnode]
        [GradNet \cite{Li2019GradNet}, wnode]
        [DIMP \cite{Bhat2019Learning}, wnode]
        [CLNet \cite{dong2020clnet}, wnode]
        [MGA-Net \cite{Wu2021Meta-Graph}, wnode]
        [RTS \cite{paul2022robust}, wnode]]
    [Continual Learning, xnode,  l sep=10mm,
        [Tracking via Continual Learning \cite{Zhang2019Long-Term}, wnode]
        [ConTACT \cite{Choi2022Visual}, wnode]]
    [Domain Adaptation, xnode,  l sep=10mm,
        [CODA \cite{Li2019Real-Time}, wnode]
        [MDNet \cite{Nam2016Learning}, wnode]]]
\end{forest}
}
    \caption{Literature Survey Tree}
    \label{fig:taxonomy}
\end{figure}
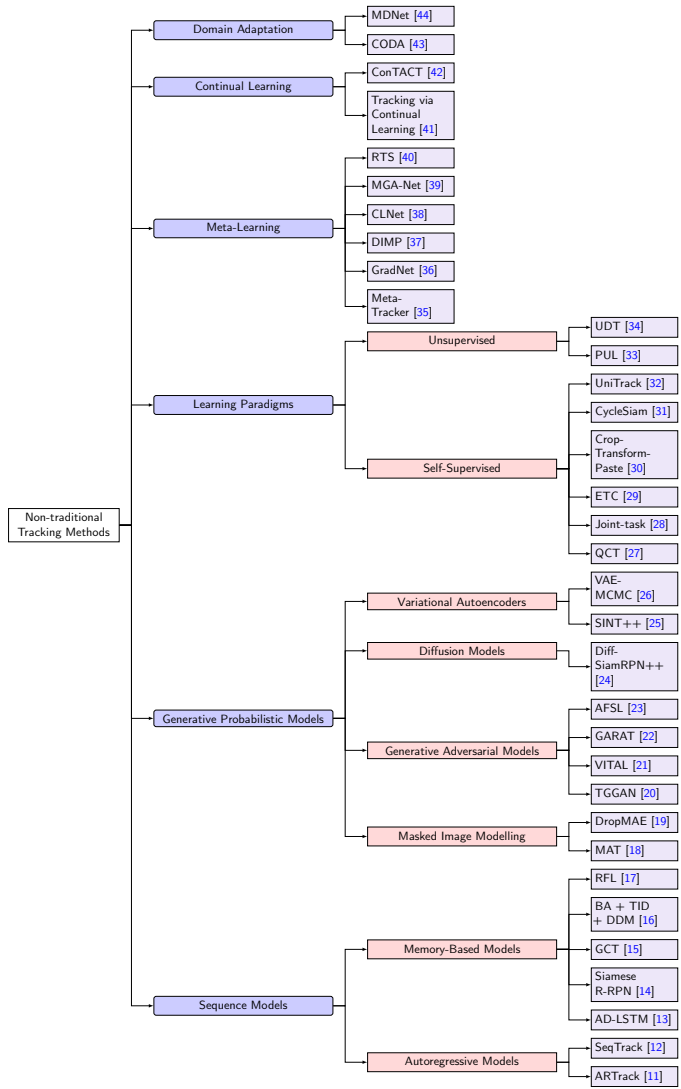

In this survey, the taxonomy mainly focuses on non-traditional visual object tracking techniques. Figure \ref{fig:taxonomy} shows the taxonomy of the surveyed methods.

\subsection{Sequence Models}
\label{sec:Sequence}

\subsubsection{Autoregressive models}
\label{sec:Autoregressive}

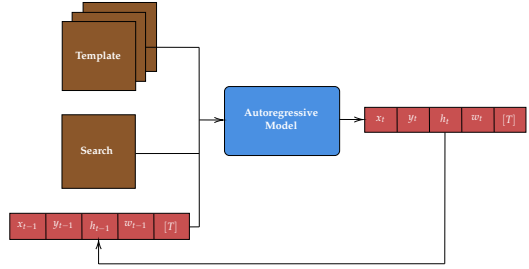
\begin{figure}[ht]
    \centering
    \tikzset{every picture/.style={line width=0.75pt}} 

    \scalebox{0.4}{ 
        \begin{tikzpicture}[x=0.75pt,y=0.75pt,yscale=-1,xscale=1]
        
        \draw  [fill={rgb, 255:red, 139; green, 87; blue, 42 }  ,fill opacity=1 ] (94.36,39.8) -- (186,39.8) -- (186,126.52) -- (94.36,126.52) -- cycle ;
        \draw  [fill={rgb, 255:red, 139; green, 87; blue, 42 }  ,fill opacity=1 ] (80.68,52.26) -- (172.32,52.26) -- (172.32,138.98) -- (80.68,138.98) -- cycle ;
        \draw  [fill={rgb, 255:red, 139; green, 87; blue, 42 }  ,fill opacity=1 ] (68.33,63.78) -- (159.97,63.78) -- (159.97,150.5) -- (68.33,150.5) -- cycle ;
        \draw  [fill={rgb, 255:red, 139; green, 87; blue, 42 }  ,fill opacity=1 ] (68,181) -- (159.5,181) -- (159.5,273.5) -- (68,273.5) -- cycle ;
        \draw  [fill={rgb, 255:red, 74; green, 144; blue, 226 }  ,fill opacity=1 ] (270.83,151.71) .. controls (270.83,148.28) and (273.61,145.5) .. (277.04,145.5) -- (408.29,145.5) .. controls (411.72,145.5) and (414.5,148.28) .. (414.5,151.71) -- (414.5,226.04) .. controls (414.5,229.47) and (411.72,232.25) .. (408.29,232.25) -- (277.04,232.25) .. controls (273.61,232.25) and (270.83,229.47) .. (270.83,226.04) -- cycle ;
        \draw  [fill={rgb, 255:red, 200; green, 80; blue, 80 }  ,fill opacity=1 ] (2.89,304.89) -- (227,304.89) -- (227,337.75) -- (2.89,337.75) -- cycle ;
        \draw [fill={rgb, 255:red, 200; green, 80; blue, 80 }  ,fill opacity=1 ]   (47.83,305.53) -- (48.01,337.43) ;
        \draw [fill={rgb, 255:red, 200; green, 80; blue, 80 }  ,fill opacity=1 ]   (92.76,305.53) -- (92.95,337.43) ;
        \draw [fill={rgb, 255:red, 200; green, 80; blue, 80 }  ,fill opacity=1 ]   (137.69,305.53) -- (137.88,337.43) ;
        \draw [fill={rgb, 255:red, 200; green, 80; blue, 80 }  ,fill opacity=1 ]   (182.63,305.53) -- (182.82,337.43) ;
        
        \draw  [fill={rgb, 255:red, 200; green, 80; blue, 80 }  ,fill opacity=1 ] (445.78,171.66) -- (647,171.66) -- (647,203) -- (445.78,203) -- cycle ;
        \draw [fill={rgb, 255:red, 200; green, 80; blue, 80 }  ,fill opacity=1 ]   (486.12,172.27) -- (486.29,202.7) ;
        \draw [fill={rgb, 255:red, 200; green, 80; blue, 80 }  ,fill opacity=1 ]   (526.47,172.27) -- (526.64,202.7) ;
        \draw [fill={rgb, 255:red, 200; green, 80; blue, 80 }  ,fill opacity=1 ]   (566.81,172.27) -- (566.98,202.7) ;
        \draw [fill={rgb, 255:red, 200; green, 80; blue, 80 }  ,fill opacity=1 ]   (607.16,172.27) -- (607.33,202.7) ;
        
        \draw    (545.95,203.2) -- (545.67,368.33) ;
        \draw    (113.67,368.44) -- (545.67,368.33) ;
        \draw    (227,322) -- (239.35,321.95) ;
        \draw    (159.95,229.75) -- (239.15,229.7) ;
        \draw [fill={rgb, 255:red, 139; green, 87; blue, 42 }  ,fill opacity=1 ]   (172.2,96.1) -- (239.15,96.25) ;
        \draw    (239.15,96.25) -- (239.35,321.95) ;
        \draw    (239.45,187.5) -- (267.1,187.59) ;
        \draw [shift={(269.1,187.6)}, rotate = 180.19] [color={rgb, 255:red, 0; green, 0; blue, 0 }  ][line width=0.75]    (10.93,-3.29) .. controls (6.95,-1.4) and (3.31,-0.3) .. (0,0) .. controls (3.31,0.3) and (6.95,1.4) .. (10.93,3.29)   ;
        \draw    (414.02,186.9) -- (441.67,186.99) ;
        \draw [shift={(443.67,187)}, rotate = 180.19] [color={rgb, 255:red, 0; green, 0; blue, 0 }  ][line width=0.75]    (10.93,-3.29) .. controls (6.95,-1.4) and (3.31,-0.3) .. (0,0) .. controls (3.31,0.3) and (6.95,1.4) .. (10.93,3.29)   ;
        \draw    (113.66,368.65) -- (113.67,341) ;
        \draw [shift={(113.67,339)}, rotate = 90.02] [color={rgb, 255:red, 0; green, 0; blue, 0 }  ][line width=0.75]    (10.93,-3.29) .. controls (6.95,-1.4) and (3.31,-0.3) .. (0,0) .. controls (3.31,0.3) and (6.95,1.4) .. (10.93,3.29)   ;
        
        \draw (10.61,314.33) node [anchor=north west][inner sep=0.75pt]   [align=left] {$\displaystyle \textcolor[rgb]{1,1,1}{x}\textcolor[rgb]{1,1,1}{_{t-1}}$};
        \draw (55.91,313.5) node [anchor=north west][inner sep=0.75pt]   [align=left] {$\displaystyle \textcolor[rgb]{1,1,1}{y}\textcolor[rgb]{1,1,1}{_{t-1}}$};
        \draw (101.41,313.5) node [anchor=north west][inner sep=0.75pt]   [align=left] {$\displaystyle \textcolor[rgb]{1,1,1}{h}\textcolor[rgb]{1,1,1}{_{t-1}}$};
        \draw (144.41,313.5) node [anchor=north west][inner sep=0.75pt]   [align=left] {$\displaystyle \textcolor[rgb]{1,1,1}{w}\textcolor[rgb]{1,1,1}{_{t-1}}$};
        \draw (193.61,314) node [anchor=north west][inner sep=0.75pt]   [align=left] {$\displaystyle \textcolor[rgb]{1,1,1}{[}\textcolor[rgb]{1,1,1}{T}\textcolor[rgb]{1,1,1}{]}$};
        \draw (458.18,180.25) node [anchor=north west][inner sep=0.75pt]   [align=left] {$\displaystyle \textcolor[rgb]{1,1,1}{x}\textcolor[rgb]{1,1,1}{_{t}}$};
        \draw (498.85,179.46) node [anchor=north west][inner sep=0.75pt]   [align=left] {$\displaystyle \textcolor[rgb]{1,1,1}{y}\textcolor[rgb]{1,1,1}{_{t}}$};
        \draw (538.81,179.46) node [anchor=north west][inner sep=0.75pt]   [align=left] {$\displaystyle \textcolor[rgb]{1,1,1}{h}\textcolor[rgb]{1,1,1}{_{t}}$};
        \draw (577.21,179.46) node [anchor=north west][inner sep=0.75pt]   [align=left] {$\displaystyle \textcolor[rgb]{1,1,1}{w}\textcolor[rgb]{1,1,1}{_{t}}$};
        \draw (615.8,178.98) node [anchor=north west][inner sep=0.75pt]   [align=left] {$\displaystyle \textcolor[rgb]{1,1,1}{[}\textcolor[rgb]{1,1,1}{T}\textcolor[rgb]{1,1,1}{]}$};
        \draw (292.85,171.51) node [anchor=north west][inner sep=0.75pt]   [align=left] {\begin{minipage}[lt]{70.75pt}\setlength\topsep{0pt}
        \begin{center}
        \textcolor[rgb]{1,1,1}{\textbf{Autoregressive}}\\\textcolor[rgb]{1,1,1}{\textbf{Model}}
        \end{center}
        
        \end{minipage}};
        \draw (89.54,218.75) node [anchor=north west][inner sep=0.75pt]   [align=left] {\textcolor[rgb]{1,1,1}{\textbf{Search}}};
        \draw (83.44,99.34) node [anchor=north west][inner sep=0.75pt]   [align=left] {\textcolor[rgb]{1,1,1}{\textbf{Template}}};

        \end{tikzpicture}

    }
    
    \caption{The autoregressive family of trackers. The template size can be different from one tracker to another. Some trackers use the full resolution of the template while others use a smaller size that focuses on the object of interest. }
    \label{fig:autoregressive_diagram}
\end{figure}

This family views single object tracking as a sequence prediction problem, just as language models do. Given an input pair of template and search images, the goal of these models is to predict a four-token sequence representing the bounding box parameters to predict the location of the template appearance in the search image as shown in Fig. \ref{fig:autoregressive_diagram}. Each of the predicted sequence tokens is causally dependent on each previous token. The training objective is to maximize the likelihood of the current token given the previous tokens and the search and template features, which is usually achieved by a cross-entropy loss function as follows:
\begin{equation}
\label{eq:cross-entropy}
    \ell = \max \sum^{T}_{t=1} \log P(Y^{t}|Y^{t-N:t-1}, X^{t}, Z)
\end{equation}
where $T$ is the number of frames. $N$ is the number of previous frames on which the generation of the current frame is dependent. $N$ is also called in literature \cite{yang2019generalized} the order of autoregression. $X^{t}$ is the search image. $Z$ is the exemplar frame, which can be the initial frame or a combination of past frames. 
Formulating the problem this way requires a vocabulary of numbers that is shared across all sequences and certainly a discrete representation of numbers so that it can be contained in a single dictionary. The implementation and architectural aspects of each model are discussed below.


\textbf{[SeqTrack]} Sequence-to-Sequence Tracker: In this paper \cite{Chen2023SeqTrack}, the model is prompted with a pair of template and search images and a box token. Each image $I\in \mathbb{R}^{3xHxW}$, is different from many works in the literature \cite{bertinetto2016fully-convolutional, ye2022joint} where the template is smaller in size than the search image. Using the full resolution makes the model more robust as the training is done with more background information. While training, the input box sequence is modelled as $[start, x, y, h, w]$. The target sequence is modelled as $[x, y, h, w, end]$ where $x, y$ are the coordinates of the bounding box center and $h, w$ is the width and height. Each continuous variable is discretized uniformly to have an integer value between $[1, n_{bins}]$ where $n_bins$ is the number of bins used to discretize the continuous parameter. The model architecture in this method is composed of an encoder-decoder architecture. The encoder is a regular vision transformer (ViT) \cite{dosovitskiy2021an} where the patches of the search and template are fed into. Half of the output patches are considered for the next stage, which is a causal decoder \cite{Vaswani2017attention}. The decoder jointly takes the encoded search and template patches and the previous bounding box token to predict the current one. During the inference phase, the prior knowledge is integrated to update the appearance model, penalizing drastic changes between two consecutive frames using a Hanning window. Moreover, the template is updated online after $T_{u}$ frames are processed according to the average probability of all tokens of the current bounding box sequence. 

\textbf{[ARTrack]} Autoregressive Visual Tracker: In this work \cite{wei2023autoregressive}, the autoregressive model is different from SeqTrack \cite{Chen2023SeqTrack} where the current sequence depends on $N$ (order of autoregression) previous sequences. This model uses different sets of inputs and outputs. First, the template size is smaller than the search. Second, the bounding box sequence representation is in the form of $[x_{min}, y_{min}, x_{max}, y_{max}]$ where $x_{min}, y_{min}$ are the coordinates of the top-left corner of the box and $x_{max}, y_{max}$ are the coordinates of the bottom-right corner. The coordinate system in this work is unified to have all coordinates of the $N$ frames referred to in one coordinate system with one global zero. Interestingly, this method does not use \textit{start} or \textit{end} tokens because the sequence size is always set to a fixed length of four. Another input is the command token, a learnable token that distinguishes between bounding box sequences when $N$ past sequences are used. The model architecture is mainly an encoder-decoder architecture. The loss function uses a combination of cross entropy and $L1$ losses.

\subsubsection{Memory-based sequence models}
\label{sec:Memory-based}

\begin{figure}[ht]
    \centering
    \tikzset{every picture/.style={line width=0.75pt}} 
    \scalebox{0.4}{
        \begin{tikzpicture}[x=0.75pt,y=0.75pt,yscale=-1,xscale=1]
        
        \draw  [fill={rgb, 255:red, 139; green, 87; blue, 42 }  ,fill opacity=1 ] (97.75,10.15) -- (189.39,10.15) -- (189.39,96.87) -- (97.75,96.87) -- cycle ;
        \draw  [fill={rgb, 255:red, 139; green, 87; blue, 42 }  ,fill opacity=1 ] (84.07,22.61) -- (175.71,22.61) -- (175.71,109.33) -- (84.07,109.33) -- cycle ;
        \draw  [fill={rgb, 255:red, 139; green, 87; blue, 42 }  ,fill opacity=1 ] (71.72,34.13) -- (163.35,34.13) -- (163.35,120.85) -- (71.72,120.85) -- cycle ;
        \draw  [fill={rgb, 255:red, 139; green, 87; blue, 42 }  ,fill opacity=1 ] (71.39,151.35) -- (162.89,151.35) -- (162.89,243.85) -- (71.39,243.85) -- cycle ;
        \draw  [fill={rgb, 255:red, 74; green, 144; blue, 226 }  ,fill opacity=1 ] (274.22,96.06) .. controls (274.22,92.63) and (277,89.85) .. (280.43,89.85) -- (411.68,89.85) .. controls (415.11,89.85) and (417.89,92.63) .. (417.89,96.06) -- (417.89,170.39) .. controls (417.89,173.82) and (415.11,176.6) .. (411.68,176.6) -- (280.43,176.6) .. controls (277,176.6) and (274.22,173.82) .. (274.22,170.39) -- cycle ;
        \draw    (242.84,131.85) -- (270.49,131.94) ;
        \draw [shift={(272.49,131.95)}, rotate = 180.19] [color={rgb, 255:red, 0; green, 0; blue, 0 }  ][line width=0.75]    (10.93,-3.29) .. controls (6.95,-1.4) and (3.31,-0.3) .. (0,0) .. controls (3.31,0.3) and (6.95,1.4) .. (10.93,3.29)   ;
        \draw    (163.34,200.1) -- (242.54,200.05) ;
        \draw [fill={rgb, 255:red, 139; green, 87; blue, 42 }  ,fill opacity=1 ]   (175.59,66.45) -- (242.54,66.6) ;
        \draw    (242.54,66.6) -- (242.54,200.05) ;
        \draw  [fill={rgb, 255:red, 74; green, 144; blue, 226 }  ,fill opacity=1 ] (273.22,212.06) .. controls (273.22,208.63) and (276,205.85) .. (279.43,205.85) -- (410.68,205.85) .. controls (414.11,205.85) and (416.89,208.63) .. (416.89,212.06) -- (416.89,286.39) .. controls (416.89,289.82) and (414.11,292.6) .. (410.68,292.6) -- (279.43,292.6) .. controls (276,292.6) and (273.22,289.82) .. (273.22,286.39) -- cycle ;
        
        \draw    (346.18,206.13) -- (346.27,178.47) ;
        \draw [shift={(346.28,176.48)}, rotate = 90.19] [color={rgb, 255:red, 0; green, 0; blue, 0 }  ][line width=0.75]    (10.93,-3.29) .. controls (6.95,-1.4) and (3.31,-0.3) .. (0,0) .. controls (3.31,0.3) and (6.95,1.4) .. (10.93,3.29)   ;
        \draw  [fill={rgb, 255:red, 74; green, 144; blue, 226 }  ,fill opacity=1 ] (448.22,212.06) .. controls (448.22,208.63) and (451,205.85) .. (454.43,205.85) -- (585.68,205.85) .. controls (589.11,205.85) and (591.89,208.63) .. (591.89,212.06) -- (591.89,286.39) .. controls (591.89,289.82) and (589.11,292.6) .. (585.68,292.6) -- (454.43,292.6) .. controls (451,292.6) and (448.22,289.82) .. (448.22,286.39) -- cycle ;
        
        \draw    (447.05,247.35) -- (419.4,247.26) ;
        \draw [shift={(417.4,247.25)}, rotate = 0.19] [color={rgb, 255:red, 0; green, 0; blue, 0 }  ][line width=0.75]    (10.93,-3.29) .. controls (6.95,-1.4) and (3.31,-0.3) .. (0,0) .. controls (3.31,0.3) and (6.95,1.4) .. (10.93,3.29)   ;
        \draw    (418.49,158.95) -- (518.68,158.96) ;
        \draw    (518.68,159.06) -- (518.74,203.46) ;
        \draw [shift={(518.74,205.46)}, rotate = 269.92] [color={rgb, 255:red, 0; green, 0; blue, 0 }  ][line width=0.75]    (10.93,-3.29) .. controls (6.95,-1.4) and (3.31,-0.3) .. (0,0) .. controls (3.31,0.3) and (6.95,1.4) .. (10.93,3.29)   ;
        \draw  [fill={rgb, 255:red, 200; green, 80; blue, 80 }  ,fill opacity=1 ] (447.58,93.66) -- (609.05,93.66) -- (609.05,125) -- (447.58,125) -- cycle ;
        \draw [fill={rgb, 255:red, 200; green, 80; blue, 80 }  ,fill opacity=1 ]   (487.93,94.27) -- (488.1,124.7) ;
        \draw [fill={rgb, 255:red, 200; green, 80; blue, 80 }  ,fill opacity=1 ]   (528.27,94.27) -- (528.44,124.7) ;
        \draw [fill={rgb, 255:red, 200; green, 80; blue, 80 }  ,fill opacity=1 ]   (568.62,94.27) -- (568.79,124.7) ;
        \draw [fill={rgb, 255:red, 200; green, 80; blue, 80 }  ,fill opacity=1 ]   (608.96,94.27) -- (609.13,124.7) ;
        \draw    (418.07,109.25) -- (445.72,109.34) ;
        \draw [shift={(447.72,109.35)}, rotate = 180.19] [color={rgb, 255:red, 0; green, 0; blue, 0 }  ][line width=0.75]    (10.93,-3.29) .. controls (6.95,-1.4) and (3.31,-0.3) .. (0,0) .. controls (3.31,0.3) and (6.95,1.4) .. (10.93,3.29)   ;
        
        \draw (92.92,189.1) node [anchor=north west][inner sep=0.75pt]   [align=left] {\textcolor[rgb]{1,1,1}{\textbf{Search}}};
        \draw (86.83,69.69) node [anchor=north west][inner sep=0.75pt]   [align=left] {\textcolor[rgb]{1,1,1}{\textbf{Template}}};
        \draw (317.23,116.86) node [anchor=north west][inner sep=0.75pt]  [color={rgb, 255:red, 255; green, 255; blue, 255 }  ,opacity=1 ] [align=left] {\begin{minipage}[lt]{41.45pt}\setlength\topsep{0pt}
        \begin{center}
        \textbf{Tracking}\\\textcolor[rgb]{1,1,1}{\textbf{Model}}
        \end{center}
        
        \end{minipage}};
        \draw (316.23,233.86) node [anchor=north west][inner sep=0.75pt]  [color={rgb, 255:red, 255; green, 255; blue, 255 }  ,opacity=1 ] [align=left] {\begin{minipage}[lt]{39.56pt}\setlength\topsep{0pt}
        \begin{center}
        \textbf{Memory}\\\textcolor[rgb]{1,1,1}{\textbf{Model}}
        \end{center}
        
        \end{minipage}};
        \draw (491.23,233.86) node [anchor=north west][inner sep=0.75pt]  [color={rgb, 255:red, 255; green, 255; blue, 255 }  ,opacity=1 ] [align=left] {\begin{minipage}[lt]{39.56pt}\setlength\topsep{0pt}
        \begin{center}
        \textbf{Memory}\\\textbf{Updater}
        \end{center}
        
        \end{minipage}};
        \draw (459.98,102.25) node [anchor=north west][inner sep=0.75pt]   [align=left] {$\displaystyle \textcolor[rgb]{1,1,1}{x}\textcolor[rgb]{1,1,1}{_{t}}$};
        \draw (500.66,101.46) node [anchor=north west][inner sep=0.75pt]   [align=left] {$\displaystyle \textcolor[rgb]{1,1,1}{y}\textcolor[rgb]{1,1,1}{_{t}}$};
        \draw (540.61,101.46) node [anchor=north west][inner sep=0.75pt]   [align=left] {$\displaystyle \textcolor[rgb]{1,1,1}{h}\textcolor[rgb]{1,1,1}{_{t}}$};
        \draw (579.02,101.46) node [anchor=north west][inner sep=0.75pt]   [align=left] {$\displaystyle \textcolor[rgb]{1,1,1}{w}\textcolor[rgb]{1,1,1}{_{t}}$};

        \end{tikzpicture}
    }
    \caption{A general diagram of memory-based sequence models.}
    \label{fig:memory_diagram}
\end{figure}
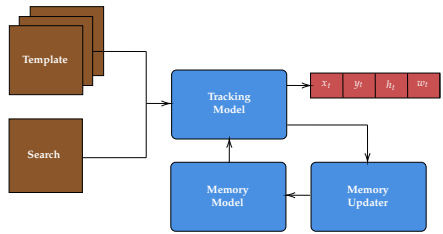

As depicted in Fig. \ref{fig:memory_diagram}, memory-based sequence models are designed to model the temporal correspondence between consecutive frames (i.e., video frames).  The appearance model in these models is formed and updated using a Recurrent Neural Network (RNN).

\textbf{[RFL]} Recurrent Filter Learning for Visual Tracking: This paper \cite{yang2017recurrent} integrated ConvLSTM \cite{shi2015convolutional} for the first time in object tracking network to exploit the spatial information captured by a CNN \cite{lecun1998gradient} and update it slowly using an LSTM as a memory unit. The network consists of an Exempler-CNN (E-CNN), which extracts the exemplar (template) features, a Search-CNN (S-CNN), which extracts the search features and a ConvLSTM module. Similar to SiamFC \cite{bertinetto2016fully-convolutional}, the goal is to get the cross-correlation between the exemplar and search features and use the correlation map to predict the object location. However, the main difference is the application of ConvLSTM on the exemplar features to recall the early exemplar states while benefitting from the current state. The loss function is the cross entropy similar to \cite{bertinetto2016fully-convolutional}.

\textbf{[GCT]} Graph Convolutional Tracking: This work \cite{gao2019graph} was the first to introduce Graph Convolutional Networks (GCNs) \cite{kipf2017semisupervised} in visual tracking. The paper proposed using a GCN to generate spatial-temporal graph embeddings of mapped historical templates $Z_{t-T:t-1}$ referred to as ST-GCN. Furthermore, the graph comprises nodes representing target parts at different time frames and edges in a fully connected manner to represent the dense dependencies between the template parts in space and throughout time. Parallel to this process, a context-aware network leverages context information from the current search image for target localization. The context embeddings are then concatenated with the spatial-temporal graph embeddings and fed into a graph learning network to form an adaptive graph. The adaptive graph is fed afterwards to a context-aware GCN that also takes the spatial-temporal graph embeddings to output an adaptive feature map. These feature maps contain information from spatial-temporal relations between template states and the context of the current search image. Finally, the adaptive feature map is cross-correlated with the search image embeddings following \cite{bertinetto2016fully-convolutional} to produce a score map, which is then utilized to locate the object in the search image.

\textbf{[Siamese R-RPN]} Cooperative Use of Recurrent Neural Network and Siamese Region Proposal Network for Robust Visual Tracking: In this work \cite{zhao2021cooperative}, the researchers built upon SiamRPN \cite{li2018high} which involves tracking via two Siamese networks for classification and regression. The idea of using Region Proposal Networks (RPNs) \cite{ren2015faster} has been used extensively in various fields in computer vision \cite{He2017mask, Kirillov2020PointRend, sun2019deep}. This is owing to its effectiveness in attending to important parts in the images where the gradient is updated following the proposal selection process. This method is intended to fix some of SiamRPN issues by introducing three new components: Channel and spatial attention, memory anchor selection and RPN integrated loss. First, channel and spatial attention is a series of convolutional layers that enhance the RPN capability to identify relevant features. Second, the memory anchor selection subnetwork is a two-layer LSTM network that takes feature maps of both the previous frame and the current frame concatenated to output and enhanced anchor based on the trajectory of the object. Henceforth, the network significantly reduces the number of region proposals and subsequently increases the performance while reducing the run time. Finally, the loss function is a combination of classification, regression, and anchor matching losses, hence ensuring that the anchor selection process is guided in a supervised manner.

\textbf{[AD-LSTM]} Attention-Driven Memory Network for Online Visual Tracking: In this method \cite{zhang2023attention}, the authors proposed AD-LSTM inspired by both long-term and short-term human visual memory in addition to a memory updater via online training. To this end, the tracker consists of four components: Feature extractor, attention-driven LSTM, convolutional short-term memory and feed-forward memory updater. First, the input search image is processed by a VGG16 network as a feature extractor \cite{Simonyan2015very}, resulting in a feature map as an output. Furthermore, the feature map is fed into the Attention-Driven \cite{hochreiter1997long} Long-Short Term Memory (AD-LSTM), which is a variation of ConvLSTM \cite{shi2015convolutional} aiming to memorize the complex appearance models of the tracked object. The different part of AD-LSTM from ConvLSTM is where spatial and channel attention to input feature maps are considered. While it is calculated independently for each channel map to obtain information about the target location and interactions with surroundings, channel attention is calculated across all channels to obtain information about the target's appearance and shape in different spaces. Even though both spatial attention and channel attention are intended for different tasks, they are implemented using mainly average and max pooling and feed-forward multi-layer perceptions (not the attention mechanism used in ViTs). Parallel to long-term memory processing, short-term memory is introduced to tackle the subtle change in the appearance of the tracked object. Both memory outputs are combined to form a response map to determine the exact target object location. After that, for upcoming frames after the first one, the tracking result is fed to a memory updater to determine whether the current tracking result can be used to update the long-term memory. The memory updater is a convolutional network that outputs a real number score according to which the current frame can be considered based on a certain threshold in the update process. Finally, a natural choice for the objective function is to minimize the mean of squared differences between the predicted response map and the ground truth, which can be considered a classification problem. The learning objective becomes trivial since the dominant labels within the response map are negative. A better objective is to minimize a gradient harmonized loss function \cite{ZHANG2022Residual}, which gives more weight to valuable (hard) samples in comparison to easy samples (negatives). The idea is intended to alleviate the effect of easy negatives by determining which samples are easy according to the L2 loss gradient at each sample and then counting the samples at each gradient value. Finally, inversely proportional weights to the number of samples assigned at each gradient value are calculated. These weights determine the importance of each sample.

\textbf{[BA+TID+DDM]} Robust Tracking via Fully Exploring Background Prior Knowledge: In this method \cite{wang2023robust}, the goal is to memorize distractors that appear in the background by explicitly modelling and mining them. First, the template and search regions are embedded using ResNet50 \cite{He2016deep}. For both regions, $X, Z \in \mathbb{R}^{3xHxW}$. After that, the object and background features are extracted from the template features to exploit the Background Area (BA). Second, a Discriminative Distractors Mining (DDM) module calculates the similarity between the target object and different background areas. The similarity (inner product) selects the top $K$ distractors to feed them with the target and search embeddings into a transformer-based encoder. Through the DDM module, prior knowledge of what parts confuse the model (i.e. distractors) can be explicitly modelled, and other redundant parts are discarded. Following the DDM, a multi-head self-attention Transformer is used to interactively mix the concatenated object, background (distractors), and search features. The self-attention mechanism forces the exploitation of useful distractor information from the template image and context information from the search image to locate the target and have high confidence in prediction by distractor awareness. Consequently, a Target-distractor Independent Decoder (TID) is presented as a memory of previous appearances of the distractors using multi-head cross attention. The memory is accessed by combining target and search features and another combination of the distractors and search features. Lastly, a classification and regression head is employed on top of the fused target-distractor-context to predict the target's confidence and location as in \cite{ye2022joint}. 


\subsection{Probabilistic Generative Models}
\label{sec:Probabilistic}

\subsubsection{Masked Image Modelling}
\label{sec:MAEs}

\begin{figure}[ht]
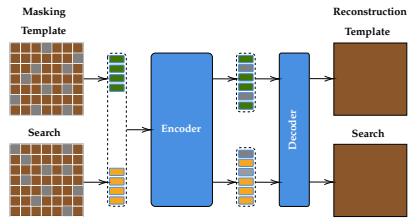

    \tikzset{every picture/.style={line width=0.75pt}} 
    \centering
    \scalebox{0.4}{

    }
    \caption{A general pertaining scheme using a masked autoencoder in single object tracking.  }
    \label{fig:mae_diagram}
\end{figure}

After a remarkable success in Natural Language Processing (NLP) \cite{kenton2019bert, lewis2020bart, Kitaev2020Reformer}, the masked information modelling paradigm made its way into the vision domain as masked image modelling \cite{he2022masked}. It also attained astonishing achievements in many vision subfields such as classification \cite{he2022masked}, semantic segmentation \cite{Wang2022ContrastMask} and image generation \cite{Chang2022MaskGIT}. In a single object tracking setting, the masked autoencoder is generally used to mask two time-dependent frames of the same sequence and predict parts or whole of them as depicted in Fig.~\ref{fig:mae_diagram}.

\textbf{[MAT]} Representation Learning for Visual Object Tracking by Masked Appearance Transfer: This method \cite{Zhao2023Representation} is an extension of \cite{he2022masked} in the visual tracking field. Both the template and search images masked patches are fed into an encoder. Two lightweight decoders that share weights are employed to reconstruct the search image and template appearance within the search region. This is done based on the assumption that reconstructing both images pushes the encoder to cluster the embeddings of the template closer to its corresponding search embeddings. Based on this method, the tracker uses the pre-trained encoder to embed the search and template images jointly. Following that, a cross-correlation operation between the search's and the template's jointly produced embeddings is performed to obtain a final representation. Based on that representation, a simple bounding box network that adapts \cite{law2018cornernet} is used to infer the corners of the bounding box. 

\textbf{[DropMAE]} Masked Autoencoders with Spatial-Attention Dropout for Tracking Tasks: This work \cite{Wu2023DropMAE} shows that the direct adaptation of MAE to the single object tracking task overlooks the temporal cues in the frames in favour of the spatial cues. The proposed solution is that, given two input frames, an adaptive spatial dropout module is applied to each head of the decoder multi-head self-attention module. This module drops out large-value attention elements in the spatial dimension with a certain probability if these elements have good matches in the temporal dimension. The spatial dropout learns better when two identity embeddings that distinguish the input frames are added. To finally integrate the model in a tracking setting, OSTrack \cite{ye2022joint} is used as a base method with the DropMAE pre-trained encoder and the identity embeddings added.

\subsubsection{Generative Adversarial Models}
\label{sec:GANs}
\begin{figure}
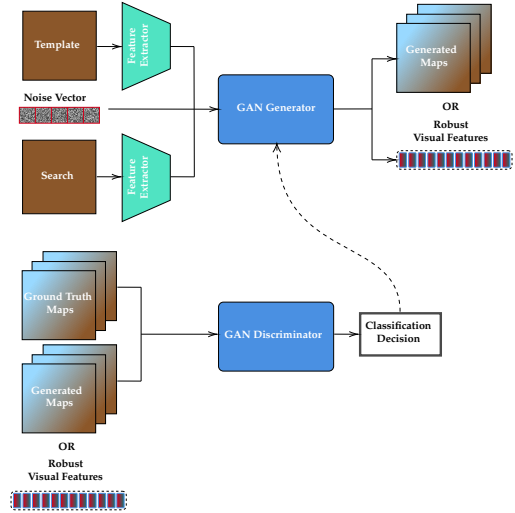

    \centering
    \tikzset {_vcptz2vcb/.code = {\pgfsetadditionalshadetransform{ \pgftransformshift{\pgfpoint{0 bp } { 0 bp }  }  \pgftransformrotate{0 }  \pgftransformscale{2 }  }}}
    \pgfdeclarehorizontalshading{_ssudrqdtq}{150bp}{rgb(0bp)=(0.82,0.01,0.11);
    rgb(37.5bp)=(0.82,0.01,0.11);
    rgb(62.5bp)=(0,0.5,0.5);
    rgb(100bp)=(0,0.5,0.5)}
    
      
    \tikzset {_3emv9w00s/.code = {\pgfsetadditionalshadetransform{ \pgftransformshift{\pgfpoint{0 bp } { 0 bp }  }  \pgftransformrotate{0 }  \pgftransformscale{2 }  }}}
    \pgfdeclarehorizontalshading{_py7site7w}{150bp}{rgb(0bp)=(0.82,0.01,0.11);
    rgb(37.5bp)=(0.82,0.01,0.11);
    rgb(62.5bp)=(0,0.5,0.5);
    rgb(100bp)=(0,0.5,0.5)}
    
      
    \tikzset {_z69bymldu/.code = {\pgfsetadditionalshadetransform{ \pgftransformshift{\pgfpoint{0 bp } { 0 bp }  }  \pgftransformrotate{0 }  \pgftransformscale{2 }  }}}
    \pgfdeclarehorizontalshading{_9yhic2jep}{150bp}{rgb(0bp)=(0.82,0.01,0.11);
    rgb(37.5bp)=(0.82,0.01,0.11);
    rgb(62.5bp)=(0,0.5,0.5);
    rgb(100bp)=(0,0.5,0.5)}
    
      
    \tikzset {_k5sikyccy/.code = {\pgfsetadditionalshadetransform{ \pgftransformshift{\pgfpoint{0 bp } { 0 bp }  }  \pgftransformrotate{0 }  \pgftransformscale{2 }  }}}
    \pgfdeclarehorizontalshading{_dw1celuws}{150bp}{rgb(0bp)=(0.82,0.01,0.11);
    rgb(37.5bp)=(0.82,0.01,0.11);
    rgb(62.5bp)=(0,0.5,0.5);
    rgb(100bp)=(0,0.5,0.5)}
    
      
    \tikzset {_ny89cup8n/.code = {\pgfsetadditionalshadetransform{ \pgftransformshift{\pgfpoint{0 bp } { 0 bp }  }  \pgftransformrotate{0 }  \pgftransformscale{2 }  }}}
    \pgfdeclarehorizontalshading{_dawf8lkjg}{150bp}{rgb(0bp)=(0.82,0.01,0.11);
    rgb(37.5bp)=(0.82,0.01,0.11);
    rgb(62.5bp)=(0,0.5,0.5);
    rgb(100bp)=(0,0.5,0.5)}
    
      
    \tikzset {_b29r4qh9q/.code = {\pgfsetadditionalshadetransform{ \pgftransformshift{\pgfpoint{0 bp } { 0 bp }  }  \pgftransformrotate{0 }  \pgftransformscale{2 }  }}}
    \pgfdeclarehorizontalshading{_0wvleamr2}{150bp}{rgb(0bp)=(0.82,0.01,0.11);
    rgb(37.5bp)=(0.82,0.01,0.11);
    rgb(62.5bp)=(0,0.5,0.5);
    rgb(100bp)=(0,0.5,0.5)}
    
      
    \tikzset {_mnqg27719/.code = {\pgfsetadditionalshadetransform{ \pgftransformshift{\pgfpoint{0 bp } { 0 bp }  }  \pgftransformrotate{0 }  \pgftransformscale{2 }  }}}
    \pgfdeclarehorizontalshading{_op728n583}{150bp}{rgb(0bp)=(0.82,0.01,0.11);
    rgb(37.5bp)=(0.82,0.01,0.11);
    rgb(62.5bp)=(0,0.5,0.5);
    rgb(100bp)=(0,0.5,0.5)}
    
      
    \tikzset {_z2rhc1bc5/.code = {\pgfsetadditionalshadetransform{ \pgftransformshift{\pgfpoint{0 bp } { 0 bp }  }  \pgftransformrotate{0 }  \pgftransformscale{2 }  }}}
    \pgfdeclarehorizontalshading{_505lnfwe4}{150bp}{rgb(0bp)=(0.82,0.01,0.11);
    rgb(37.5bp)=(0.82,0.01,0.11);
    rgb(62.5bp)=(0,0.5,0.5);
    rgb(100bp)=(0,0.5,0.5)}
    
      
    \tikzset {_6z64oundt/.code = {\pgfsetadditionalshadetransform{ \pgftransformshift{\pgfpoint{0 bp } { 0 bp }  }  \pgftransformrotate{0 }  \pgftransformscale{2 }  }}}
    \pgfdeclarehorizontalshading{_ugg2hokpe}{150bp}{rgb(0bp)=(0.82,0.01,0.11);
    rgb(37.5bp)=(0.82,0.01,0.11);
    rgb(62.5bp)=(0,0.5,0.5);
    rgb(100bp)=(0,0.5,0.5)}
    
      
    \tikzset {_dn0xzi87p/.code = {\pgfsetadditionalshadetransform{ \pgftransformshift{\pgfpoint{0 bp } { 0 bp }  }  \pgftransformrotate{0 }  \pgftransformscale{2 }  }}}
    \pgfdeclarehorizontalshading{_r69dmtzu1}{150bp}{rgb(0bp)=(0.82,0.01,0.11);
    rgb(37.5bp)=(0.82,0.01,0.11);
    rgb(62.5bp)=(0,0.5,0.5);
    rgb(100bp)=(0,0.5,0.5)}
    
      
    \tikzset {_i28bb72jx/.code = {\pgfsetadditionalshadetransform{ \pgftransformshift{\pgfpoint{0 bp } { 0 bp }  }  \pgftransformrotate{0 }  \pgftransformscale{2 }  }}}
    \pgfdeclarehorizontalshading{_i1iqo2rxn}{150bp}{rgb(0bp)=(0.82,0.01,0.11);
    rgb(37.5bp)=(0.82,0.01,0.11);
    rgb(62.5bp)=(0,0.5,0.5);
    rgb(100bp)=(0,0.5,0.5)}
    
      
    \tikzset {_zmwokolnk/.code = {\pgfsetadditionalshadetransform{ \pgftransformshift{\pgfpoint{0 bp } { 0 bp }  }  \pgftransformrotate{0 }  \pgftransformscale{2 }  }}}
    \pgfdeclarehorizontalshading{_9m0hfvqi5}{150bp}{rgb(0bp)=(0.82,0.01,0.11);
    rgb(37.5bp)=(0.82,0.01,0.11);
    rgb(62.5bp)=(0,0.5,0.5);
    rgb(100bp)=(0,0.5,0.5)}
    
      
    \tikzset {_7o6cywd96/.code = {\pgfsetadditionalshadetransform{ \pgftransformshift{\pgfpoint{0 bp } { 0 bp }  }  \pgftransformrotate{-35 }  \pgftransformscale{2 }  }}}
    \pgfdeclarehorizontalshading{_gyx39ue4n}{150bp}{rgb(0bp)=(0.6,0.85,1);
    rgb(37.5bp)=(0.6,0.85,1);
    rgb(62.5bp)=(0.55,0.34,0.16);
    rgb(100bp)=(0.55,0.34,0.16)}
    
      
    \tikzset {_or317b2wi/.code = {\pgfsetadditionalshadetransform{ \pgftransformshift{\pgfpoint{0 bp } { 0 bp }  }  \pgftransformrotate{-35 }  \pgftransformscale{2 }  }}}
    \pgfdeclarehorizontalshading{_dpqcdf81k}{150bp}{rgb(0bp)=(0.6,0.85,1);
    rgb(37.5bp)=(0.6,0.85,1);
    rgb(62.5bp)=(0.55,0.34,0.16);
    rgb(100bp)=(0.55,0.34,0.16)}
    
      
    \tikzset {_jrlnp2acw/.code = {\pgfsetadditionalshadetransform{ \pgftransformshift{\pgfpoint{0 bp } { 0 bp }  }  \pgftransformrotate{-35 }  \pgftransformscale{2 }  }}}
    \pgfdeclarehorizontalshading{_fmpt4cucy}{150bp}{rgb(0bp)=(0.6,0.85,1);
    rgb(37.5bp)=(0.6,0.85,1);
    rgb(62.5bp)=(0.55,0.34,0.16);
    rgb(100bp)=(0.55,0.34,0.16)}
    
      
    \tikzset {_nwsy44jzi/.code = {\pgfsetadditionalshadetransform{ \pgftransformshift{\pgfpoint{0 bp } { 0 bp }  }  \pgftransformrotate{0 }  \pgftransformscale{2 }  }}}
    \pgfdeclarehorizontalshading{_kjqnhipol}{150bp}{rgb(0bp)=(0.82,0.01,0.11);
    rgb(37.5bp)=(0.82,0.01,0.11);
    rgb(62.5bp)=(0,0.5,0.5);
    rgb(100bp)=(0,0.5,0.5)}
    
      
    \tikzset {_0891vks2q/.code = {\pgfsetadditionalshadetransform{ \pgftransformshift{\pgfpoint{0 bp } { 0 bp }  }  \pgftransformrotate{0 }  \pgftransformscale{2 }  }}}
    \pgfdeclarehorizontalshading{_m3s0fouwa}{150bp}{rgb(0bp)=(0.82,0.01,0.11);
    rgb(37.5bp)=(0.82,0.01,0.11);
    rgb(62.5bp)=(0,0.5,0.5);
    rgb(100bp)=(0,0.5,0.5)}
    
      
    \tikzset {_lqgddcyre/.code = {\pgfsetadditionalshadetransform{ \pgftransformshift{\pgfpoint{0 bp } { 0 bp }  }  \pgftransformrotate{0 }  \pgftransformscale{2 }  }}}
    \pgfdeclarehorizontalshading{_yudotjxg7}{150bp}{rgb(0bp)=(0.82,0.01,0.11);
    rgb(37.5bp)=(0.82,0.01,0.11);
    rgb(62.5bp)=(0,0.5,0.5);
    rgb(100bp)=(0,0.5,0.5)}
    
      
    \tikzset {_dq01faegp/.code = {\pgfsetadditionalshadetransform{ \pgftransformshift{\pgfpoint{0 bp } { 0 bp }  }  \pgftransformrotate{0 }  \pgftransformscale{2 }  }}}
    \pgfdeclarehorizontalshading{_lp7o6ggjr}{150bp}{rgb(0bp)=(0.82,0.01,0.11);
    rgb(37.5bp)=(0.82,0.01,0.11);
    rgb(62.5bp)=(0,0.5,0.5);
    rgb(100bp)=(0,0.5,0.5)}
    
      
    \tikzset {_oumoj275t/.code = {\pgfsetadditionalshadetransform{ \pgftransformshift{\pgfpoint{0 bp } { 0 bp }  }  \pgftransformrotate{0 }  \pgftransformscale{2 }  }}}
    \pgfdeclarehorizontalshading{_hqj0mlerl}{150bp}{rgb(0bp)=(0.82,0.01,0.11);
    rgb(37.5bp)=(0.82,0.01,0.11);
    rgb(62.5bp)=(0,0.5,0.5);
    rgb(100bp)=(0,0.5,0.5)}
    
      
    \tikzset {_gvgzeycl0/.code = {\pgfsetadditionalshadetransform{ \pgftransformshift{\pgfpoint{0 bp } { 0 bp }  }  \pgftransformrotate{0 }  \pgftransformscale{2 }  }}}
    \pgfdeclarehorizontalshading{_mbd6zhkyl}{150bp}{rgb(0bp)=(0.82,0.01,0.11);
    rgb(37.5bp)=(0.82,0.01,0.11);
    rgb(62.5bp)=(0,0.5,0.5);
    rgb(100bp)=(0,0.5,0.5)}
    
      
    \tikzset {_ybqt5euza/.code = {\pgfsetadditionalshadetransform{ \pgftransformshift{\pgfpoint{0 bp } { 0 bp }  }  \pgftransformrotate{0 }  \pgftransformscale{2 }  }}}
    \pgfdeclarehorizontalshading{_bi8mpynv1}{150bp}{rgb(0bp)=(0.82,0.01,0.11);
    rgb(37.5bp)=(0.82,0.01,0.11);
    rgb(62.5bp)=(0,0.5,0.5);
    rgb(100bp)=(0,0.5,0.5)}
    
      
    \tikzset {_c9iasr6oa/.code = {\pgfsetadditionalshadetransform{ \pgftransformshift{\pgfpoint{0 bp } { 0 bp }  }  \pgftransformrotate{0 }  \pgftransformscale{2 }  }}}
    \pgfdeclarehorizontalshading{_5mzti30e9}{150bp}{rgb(0bp)=(0.82,0.01,0.11);
    rgb(37.5bp)=(0.82,0.01,0.11);
    rgb(62.5bp)=(0,0.5,0.5);
    rgb(100bp)=(0,0.5,0.5)}
    
      
    \tikzset {_1q3mqf9ro/.code = {\pgfsetadditionalshadetransform{ \pgftransformshift{\pgfpoint{0 bp } { 0 bp }  }  \pgftransformrotate{0 }  \pgftransformscale{2 }  }}}
    \pgfdeclarehorizontalshading{_sqipcbpaf}{150bp}{rgb(0bp)=(0.82,0.01,0.11);
    rgb(37.5bp)=(0.82,0.01,0.11);
    rgb(62.5bp)=(0,0.5,0.5);
    rgb(100bp)=(0,0.5,0.5)}
    
      
    \tikzset {_277x2eugn/.code = {\pgfsetadditionalshadetransform{ \pgftransformshift{\pgfpoint{0 bp } { 0 bp }  }  \pgftransformrotate{0 }  \pgftransformscale{2 }  }}}
    \pgfdeclarehorizontalshading{_engyc1ttl}{150bp}{rgb(0bp)=(0.82,0.01,0.11);
    rgb(37.5bp)=(0.82,0.01,0.11);
    rgb(62.5bp)=(0,0.5,0.5);
    rgb(100bp)=(0,0.5,0.5)}
    
      
    \tikzset {_p6hhhqfp6/.code = {\pgfsetadditionalshadetransform{ \pgftransformshift{\pgfpoint{0 bp } { 0 bp }  }  \pgftransformrotate{0 }  \pgftransformscale{2 }  }}}
    \pgfdeclarehorizontalshading{_wgkmfo11m}{150bp}{rgb(0bp)=(0.82,0.01,0.11);
    rgb(37.5bp)=(0.82,0.01,0.11);
    rgb(62.5bp)=(0,0.5,0.5);
    rgb(100bp)=(0,0.5,0.5)}
    
      
    \tikzset {_b1edyczds/.code = {\pgfsetadditionalshadetransform{ \pgftransformshift{\pgfpoint{0 bp } { 0 bp }  }  \pgftransformrotate{0 }  \pgftransformscale{2 }  }}}
    \pgfdeclarehorizontalshading{_w7olqm9uq}{150bp}{rgb(0bp)=(0.82,0.01,0.11);
    rgb(37.5bp)=(0.82,0.01,0.11);
    rgb(62.5bp)=(0,0.5,0.5);
    rgb(100bp)=(0,0.5,0.5)}
    
      
    \tikzset {_ix8dr70kn/.code = {\pgfsetadditionalshadetransform{ \pgftransformshift{\pgfpoint{0 bp } { 0 bp }  }  \pgftransformrotate{-35 }  \pgftransformscale{2 }  }}}
    \pgfdeclarehorizontalshading{_lx98r9wxt}{150bp}{rgb(0bp)=(0.6,0.85,1);
    rgb(37.5bp)=(0.6,0.85,1);
    rgb(62.5bp)=(0.55,0.34,0.16);
    rgb(100bp)=(0.55,0.34,0.16)}
    
      
    \tikzset {_f8j35ay4e/.code = {\pgfsetadditionalshadetransform{ \pgftransformshift{\pgfpoint{0 bp } { 0 bp }  }  \pgftransformrotate{-35 }  \pgftransformscale{2 }  }}}
    \pgfdeclarehorizontalshading{_wk3ykhgyc}{150bp}{rgb(0bp)=(0.6,0.85,1);
    rgb(37.5bp)=(0.6,0.85,1);
    rgb(62.5bp)=(0.55,0.34,0.16);
    rgb(100bp)=(0.55,0.34,0.16)}
    
      
    \tikzset {_zwyllobdj/.code = {\pgfsetadditionalshadetransform{ \pgftransformshift{\pgfpoint{0 bp } { 0 bp }  }  \pgftransformrotate{-35 }  \pgftransformscale{2 }  }}}
    \pgfdeclarehorizontalshading{_nowij89w8}{150bp}{rgb(0bp)=(0.6,0.85,1);
    rgb(37.5bp)=(0.6,0.85,1);
    rgb(62.5bp)=(0.55,0.34,0.16);
    rgb(100bp)=(0.55,0.34,0.16)}
    
      
    \tikzset {_2b9l6eoyh/.code = {\pgfsetadditionalshadetransform{ \pgftransformshift{\pgfpoint{0 bp } { 0 bp }  }  \pgftransformrotate{-35 }  \pgftransformscale{2 }  }}}
    \pgfdeclarehorizontalshading{_micbr1hzf}{150bp}{rgb(0bp)=(0.6,0.85,1);
    rgb(37.5bp)=(0.6,0.85,1);
    rgb(62.5bp)=(0.55,0.34,0.16);
    rgb(100bp)=(0.55,0.34,0.16)}
    
      
    \tikzset {_kh37napjc/.code = {\pgfsetadditionalshadetransform{ \pgftransformshift{\pgfpoint{0 bp } { 0 bp }  }  \pgftransformrotate{-35 }  \pgftransformscale{2 }  }}}
    \pgfdeclarehorizontalshading{_9wfrhsrmi}{150bp}{rgb(0bp)=(0.6,0.85,1);
    rgb(37.5bp)=(0.6,0.85,1);
    rgb(62.5bp)=(0.55,0.34,0.16);
    rgb(100bp)=(0.55,0.34,0.16)}
    
      
    \tikzset {_lb46h74hr/.code = {\pgfsetadditionalshadetransform{ \pgftransformshift{\pgfpoint{0 bp } { 0 bp }  }  \pgftransformrotate{-35 }  \pgftransformscale{2 }  }}}
    \pgfdeclarehorizontalshading{_stjmw2hwa}{150bp}{rgb(0bp)=(0.6,0.85,1);
    rgb(37.5bp)=(0.6,0.85,1);
    rgb(62.5bp)=(0.55,0.34,0.16);
    rgb(100bp)=(0.55,0.34,0.16)}
    \tikzset{every picture/.style={line width=0.75pt}} 
    \scalebox{0.4}{

        
    }
    \caption{A general diagram of GAN-based single object tracking methods. The input to the generator is the features of the template image or a combination of the features of the template and the search. The generator output is either a map representing an enhanced set of template appearances or a robust feature representation depending on the model design. Either output is fed to a discriminator alongside ground truth maps that represent parts of the search image or the template itself. The discriminator mainly outputs a classification decision reflecting a final score map or a scaler that determines the quality of a generator output.}
    \label{fig:gan_diagram}
\end{figure}

Generative Adversarial Networks (GANs) \cite{Goodfellow2014Generative} have greatly impacted computer vision society since their introduction. GANs are still widely used in many subfields such as image generation \cite{Liu2020Diverse, Wang2018Transferring, han2018GAN-Based}, object detection \cite{Mustikovela2021Self-Supervised, Li2017Perceptual} and semantic segmentation \cite{Souly2017Semi, Zhang2021Multitask} indicating their effectiveness. Conditional GAN typically involves a minimax game mechanism where the generator objective is to minimize the log-likelihood while the discriminator objective is to maximize the probability of assigning the correct label as follows: 
\begin{equation}
\label{eq:GAN}
    \ell = \mathbf{E}_{x, y \sim p_{data}(x, y)}[\log(D(x, y))] + \mathbf{E}_{x \sim p_{data}(x), z \sim p_{z}(z)}[\log(1-D(x, G(x, z)))]
\end{equation}
where $G$ is the generator that learns to map data $x$ from distribution $p_{data}(x)$ and $z$ from noise distribution $p_{z}(z)$ to the distribution over $y$, and $D$ represents the discriminator. A general scheme of GANs in single object tracking is shown in Fig.~\ref{fig:gan_diagram}.

\textbf{[TGGAN]} Generating Reliable Online Adaptive Templates for Visual Tracking: This method \cite{Guo2018generating} enhances Siamese tracking by generating diverse target object views. Guided by ground truth templates, a generator creates templates simulating object appearance changes while a discriminator evaluates their quality. The network undergoes two phases: first, a Siamese-based CNN is trained on tracking data, then the generator and discriminator are trained to meet GAN and Task-Guided (TG) objectives. The Siamese network remains fixed during GAN training and operates similarly to standard tracking. The discriminator learns to classify authentic from fake images, with the loss function combining GAN and margin contrastive losses for the Siamese branch output. During tracking, the Siamese network uses the enhanced set of generated exemplars alongside the current search image to produce a similarity map. The total loss function combines the GAN loss and a margin contrastive loss \cite{Chopra2005learning} for the Siamese branch output.


\textbf{[VITAL]} Visual Tracking via Adversarial Learning: This work \cite{Song2018VITAL} proposes an adversarial training of an existing tracking-by-detection framework following the literature \cite{li2018high, ning2016object}. For any tracking-by-detection technique, a detector is applied first, and then a classifier is used to discriminate between the positive target and negative background. This method introduces a generator as an intermediate step while the classifier is viewed as the GAN discriminator. Flowing in the network during training, a search image gets its features extracted. The features are then fed into the generator. The generator produces a mask, initialized by random noise, that drops out some search features to degrade the discriminator the most. If the generator can do so, then the discriminator has learned the discriminative features of the target and implicitly how object distractors may appear. It is crucial to notice that the loss function adopted in this method cures the imbalance between the positive and negative samples because most of the classification map is background (easy negative). The curation is done via a modulation factor that increases the weight of the positive samples. During tracking time, the generator is removed, and the discriminator acts like a classifier in a regular tracking-by-detection mechanism.

\textbf{[AFSL]} Adversarial Feature Sampling Learning for Efficient Visual Tracking: In a typical tracking-by-detection setting, many region proposals are sampled from the search image using an area around the target location in the previous frame. This method \cite{Yin2020Adversarial}, proposes sampling in the feature space using Spatial Bilinear Resampling (SBR) \cite{Ariawan2008Nearest-neighbor} which mainly samples region proposals from the feature maps obtained from the feature extractor. Although sampling from embedded feature maps is beneficial regarding FPS speed, it involves many overlapping features in the feature space. The authors propose sampling from two different feature spaces and fusing the sampled features. This way, the sampled features capture semantic relations and target appearance variation. A GAN in this method mitigates the effect of overlap between samples by dropping out some samples by leveraging the adaptive dropout algorithm provided by VITAL \cite{Song2018VITAL}. The GAN generator selects the mask with the largest loss to feed into the discriminator to enhance discriminative feature detection.

\textbf{[GARAT]} Generative Adversarial Learning for Robust and Accurate Tracking: This work \cite{Yao2022GARAT} enhances the Siamese tracking pipeline by removing distractors in the search image. The network comprises a Siamese CNN \cite{bertinetto2016fully-convolutional} and a GAN. During training, the GAN is trained first to enable its generator to produce a set of distractors that mimic the appearance change of the target object. Distractor ground truth images are obtained from the training set by subtracting the ground truth target from the search image. During GAN training, the GAN discriminator receives the generated distractors and the ground truth distractors to output a classification score indicating the quality of the generated distractors. Consequently, the Siamese network is trained on search images without the distractors by subtracting the search image from the distractors generated by the generator. During tracking, the generator, the only GAN part used in this phase, cancels the distractors from the search image. Finally, a distractor-free image is fed to the Siamese network. 

\subsubsection{Diffusion Models}
\label{sec:Diffusion}

Since their introduction \cite{Jascha2015Deep}, thermodynamics-inspired diffusion models have impacted the computer vision society. Based on this idea, denoising diffusion models were introduced \cite{Ho2020Denoising} to mark the start of a new era in image generation. Henceforth, denoising diffusion models have been ubiquitous in many computer vision applications such as deep image generation \cite{Ho2022Cascaded, Dhariwal2021Diffusion, Gu2022Vector}, semantic segmentation \cite{Peng2023Diffusion-based, Yang2023Diffusion}, object detection \cite{Chen2023DiffusionDet} and recently multi-object tracking \cite{luo2023diffusiontrack}.

\textbf{[Diff-SiamRPN]} Siamese Network for Object Tracking with Diffusion Model: This work \cite{Zhang2023Siamese} integrates the concept of Denoising Diffusion Implicit Models (DDIM) \cite{Song2021Denoising} into SiamRPN \cite{li2018high}. Firstly, a DDIM model generates $n$ images prompted by the template image. These images are assumed to enrich the object appearance modelling by integrating many scenarios in which the target object can be, such as different lighting conditions or multiple poses. Finally, the SiamRPN network is trained using the generated images as templates by fusing the $n$ templates in one template map. Notably, the DDIM is only trained on the general-purpose ImageNet-1K dataset \cite{Deng2009ImageNet} and is prompted with template images from single object tracking datasets such as GOT-10K \cite{Huang2021GOT-10K} and LASOT \cite{Fan2019LaSOT}.

\subsubsection{Variational Autoencoders}
\label{sec:VAEs}

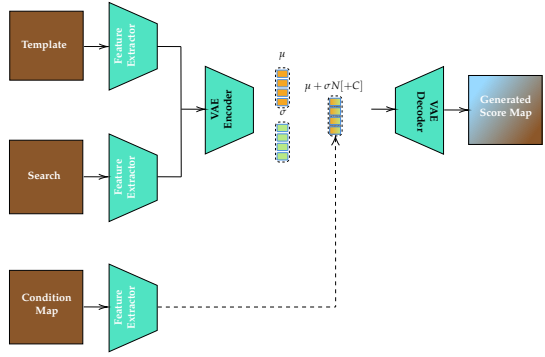
\begin{figure}
    \centering
  
    \tikzset {_pyq9s8tbr/.code = {\pgfsetadditionalshadetransform{ \pgftransformshift{\pgfpoint{0 bp } { 0 bp }  }  \pgftransformrotate{0 }  \pgftransformscale{2 }  }}}
    \pgfdeclarehorizontalshading{_syfd6upif}{150bp}{rgb(0bp)=(0.96,0.65,0.14);
    rgb(37.5bp)=(0.96,0.65,0.14);
    rgb(62.5bp)=(0.72,0.91,0.53);
    rgb(100bp)=(0.72,0.91,0.53)}
    
      
    \tikzset {_iah63mk04/.code = {\pgfsetadditionalshadetransform{ \pgftransformshift{\pgfpoint{0 bp } { 0 bp }  }  \pgftransformrotate{0 }  \pgftransformscale{2 }  }}}
    \pgfdeclarehorizontalshading{_yvoidsw50}{150bp}{rgb(0bp)=(0.96,0.65,0.14);
    rgb(37.5bp)=(0.96,0.65,0.14);
    rgb(62.5bp)=(0.72,0.91,0.53);
    rgb(100bp)=(0.72,0.91,0.53)}
    
      
    \tikzset {_p8w5puypp/.code = {\pgfsetadditionalshadetransform{ \pgftransformshift{\pgfpoint{0 bp } { 0 bp }  }  \pgftransformrotate{0 }  \pgftransformscale{2 }  }}}
    \pgfdeclarehorizontalshading{_281ocr85x}{150bp}{rgb(0bp)=(0.96,0.65,0.14);
    rgb(37.5bp)=(0.96,0.65,0.14);
    rgb(62.5bp)=(0.72,0.91,0.53);
    rgb(100bp)=(0.72,0.91,0.53)}
    
      
    \tikzset {_ji8tb49th/.code = {\pgfsetadditionalshadetransform{ \pgftransformshift{\pgfpoint{0 bp } { 0 bp }  }  \pgftransformrotate{0 }  \pgftransformscale{2 }  }}}
    \pgfdeclarehorizontalshading{_x2ut3z3mr}{150bp}{rgb(0bp)=(0.96,0.65,0.14);
    rgb(37.5bp)=(0.96,0.65,0.14);
    rgb(62.5bp)=(0.72,0.91,0.53);
    rgb(100bp)=(0.72,0.91,0.53)}
    
      
    \tikzset {_yh88qqgdj/.code = {\pgfsetadditionalshadetransform{ \pgftransformshift{\pgfpoint{0 bp } { 0 bp }  }  \pgftransformrotate{-35 }  \pgftransformscale{2 }  }}}
    \pgfdeclarehorizontalshading{_5xd7kubg7}{150bp}{rgb(0bp)=(0.6,0.85,1);
    rgb(37.5bp)=(0.6,0.85,1);
    rgb(62.5bp)=(0.55,0.34,0.16);
    rgb(100bp)=(0.55,0.34,0.16)}
    \tikzset{every picture/.style={line width=0.75pt}} 
    \scalebox{0.4}{
        \begin{tikzpicture}[x=0.75pt,y=0.75pt,yscale=-1,xscale=1]
        
        \draw (414.8,158.09) node  {\includegraphics[width=10.11pt,height=33.81pt]{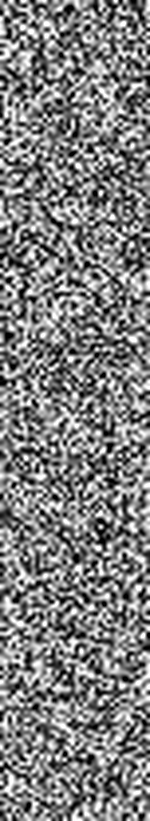}};
        \draw  [dash pattern={on 2.25pt off 2.25pt on 2.25pt off 2.25pt}] (406.16,137.68) .. controls (406.16,134.92) and (408.4,132.68) .. (411.16,132.68) -- (418.45,132.68) .. controls (421.21,132.68) and (423.45,134.92) .. (423.45,137.68) -- (423.45,177.97) .. controls (423.45,180.74) and (421.21,182.97) .. (418.45,182.97) -- (411.16,182.97) .. controls (408.4,182.97) and (406.16,180.74) .. (406.16,177.97) -- cycle ;
        \path  [shading=_syfd6upif,_pyq9s8tbr] (408.07,136.55) .. controls (408.07,135.99) and (408.52,135.55) .. (409.07,135.55) -- (420.53,135.55) .. controls (421.09,135.55) and (421.53,135.99) .. (421.53,136.55) -- (421.53,143.71) .. controls (421.53,144.27) and (421.09,144.71) .. (420.53,144.71) -- (409.07,144.71) .. controls (408.52,144.71) and (408.07,144.27) .. (408.07,143.71) -- cycle ; 
         \draw  [color={rgb, 255:red, 74; green, 144; blue, 226 }  ,draw opacity=1 ] (408.07,136.55) .. controls (408.07,135.99) and (408.52,135.55) .. (409.07,135.55) -- (420.53,135.55) .. controls (421.09,135.55) and (421.53,135.99) .. (421.53,136.55) -- (421.53,143.71) .. controls (421.53,144.27) and (421.09,144.71) .. (420.53,144.71) -- (409.07,144.71) .. controls (408.52,144.71) and (408.07,144.27) .. (408.07,143.71) -- cycle ; 
        
        \path  [shading=_yvoidsw50,_iah63mk04] (408.07,172.55) .. controls (408.07,171.99) and (408.52,171.55) .. (409.07,171.55) -- (420.53,171.55) .. controls (421.09,171.55) and (421.53,171.99) .. (421.53,172.55) -- (421.53,179.71) .. controls (421.53,180.27) and (421.09,180.71) .. (420.53,180.71) -- (409.07,180.71) .. controls (408.52,180.71) and (408.07,180.27) .. (408.07,179.71) -- cycle ; 
         \draw  [color={rgb, 255:red, 74; green, 144; blue, 226 }  ,draw opacity=1 ] (408.07,172.55) .. controls (408.07,171.99) and (408.52,171.55) .. (409.07,171.55) -- (420.53,171.55) .. controls (421.09,171.55) and (421.53,171.99) .. (421.53,172.55) -- (421.53,179.71) .. controls (421.53,180.27) and (421.09,180.71) .. (420.53,180.71) -- (409.07,180.71) .. controls (408.52,180.71) and (408.07,180.27) .. (408.07,179.71) -- cycle ; 
        
        \path  [shading=_281ocr85x,_p8w5puypp] (408.07,160.55) .. controls (408.07,159.99) and (408.52,159.55) .. (409.07,159.55) -- (420.53,159.55) .. controls (421.09,159.55) and (421.53,159.99) .. (421.53,160.55) -- (421.53,167.71) .. controls (421.53,168.27) and (421.09,168.71) .. (420.53,168.71) -- (409.07,168.71) .. controls (408.52,168.71) and (408.07,168.27) .. (408.07,167.71) -- cycle ; 
         \draw  [color={rgb, 255:red, 74; green, 144; blue, 226 }  ,draw opacity=1 ] (408.07,160.55) .. controls (408.07,159.99) and (408.52,159.55) .. (409.07,159.55) -- (420.53,159.55) .. controls (421.09,159.55) and (421.53,159.99) .. (421.53,160.55) -- (421.53,167.71) .. controls (421.53,168.27) and (421.09,168.71) .. (420.53,168.71) -- (409.07,168.71) .. controls (408.52,168.71) and (408.07,168.27) .. (408.07,167.71) -- cycle ; 
        
        \path  [shading=_x2ut3z3mr,_ji8tb49th] (408.07,148.55) .. controls (408.07,147.99) and (408.52,147.55) .. (409.07,147.55) -- (420.53,147.55) .. controls (421.09,147.55) and (421.53,147.99) .. (421.53,148.55) -- (421.53,155.71) .. controls (421.53,156.27) and (421.09,156.71) .. (420.53,156.71) -- (409.07,156.71) .. controls (408.52,156.71) and (408.07,156.27) .. (408.07,155.71) -- cycle ; 
         \draw  [color={rgb, 255:red, 74; green, 144; blue, 226 }  ,draw opacity=1 ] (408.07,148.55) .. controls (408.07,147.99) and (408.52,147.55) .. (409.07,147.55) -- (420.53,147.55) .. controls (421.09,147.55) and (421.53,147.99) .. (421.53,148.55) -- (421.53,155.71) .. controls (421.53,156.27) and (421.09,156.71) .. (420.53,156.71) -- (409.07,156.71) .. controls (408.52,156.71) and (408.07,156.27) .. (408.07,155.71) -- cycle ; 

        \draw  [dash pattern={on 2.25pt off 2.25pt on 2.25pt off 2.25pt}] (340.41,102.33) .. controls (340.41,99.57) and (342.65,97.33) .. (345.41,97.33) -- (352.7,97.33) .. controls (355.46,97.33) and (357.7,99.57) .. (357.7,102.33) -- (357.7,142.62) .. controls (357.7,145.39) and (355.46,147.62) .. (352.7,147.62) -- (345.41,147.62) .. controls (342.65,147.62) and (340.41,145.39) .. (340.41,142.62) -- cycle ;
        \draw  [color={rgb, 255:red, 74; green, 144; blue, 226 }  ,draw opacity=1 ][fill={rgb, 255:red, 245; green, 166; blue, 35 }  ,fill opacity=1 ] (342.32,101.2) .. controls (342.32,100.64) and (342.77,100.2) .. (343.32,100.2) -- (354.78,100.2) .. controls (355.34,100.2) and (355.78,100.64) .. (355.78,101.2) -- (355.78,108.36) .. controls (355.78,108.92) and (355.34,109.36) .. (354.78,109.36) -- (343.32,109.36) .. controls (342.77,109.36) and (342.32,108.92) .. (342.32,108.36) -- cycle ;
        \draw  [color={rgb, 255:red, 74; green, 144; blue, 226 }  ,draw opacity=1 ][fill={rgb, 255:red, 245; green, 166; blue, 35 }  ,fill opacity=1 ] (342.32,113.2) .. controls (342.32,112.64) and (342.77,112.2) .. (343.32,112.2) -- (354.78,112.2) .. controls (355.34,112.2) and (355.78,112.64) .. (355.78,113.2) -- (355.78,120.36) .. controls (355.78,120.92) and (355.34,121.36) .. (354.78,121.36) -- (343.32,121.36) .. controls (342.77,121.36) and (342.32,120.92) .. (342.32,120.36) -- cycle ;
        \draw  [color={rgb, 255:red, 74; green, 144; blue, 226 }  ,draw opacity=1 ][fill={rgb, 255:red, 245; green, 166; blue, 35 }  ,fill opacity=1 ] (342.32,125.03) .. controls (342.32,124.48) and (342.77,124.03) .. (343.32,124.03) -- (354.78,124.03) .. controls (355.34,124.03) and (355.78,124.48) .. (355.78,125.03) -- (355.78,132.2) .. controls (355.78,132.75) and (355.34,133.2) .. (354.78,133.2) -- (343.32,133.2) .. controls (342.77,133.2) and (342.32,132.75) .. (342.32,132.2) -- cycle ;
        \draw  [color={rgb, 255:red, 74; green, 144; blue, 226 }  ,draw opacity=1 ][fill={rgb, 255:red, 245; green, 166; blue, 35 }  ,fill opacity=1 ] (342.32,136.86) .. controls (342.32,136.31) and (342.77,135.86) .. (343.32,135.86) -- (354.78,135.86) .. controls (355.34,135.86) and (355.78,136.31) .. (355.78,136.86) -- (355.78,144.03) .. controls (355.78,144.58) and (355.34,145.03) .. (354.78,145.03) -- (343.32,145.03) .. controls (342.77,145.03) and (342.32,144.58) .. (342.32,144.03) -- cycle ;
        
        \draw  [dash pattern={on 2.25pt off 2.25pt on 2.25pt off 2.25pt}] (340.41,170.5) .. controls (340.41,167.74) and (342.65,165.5) .. (345.41,165.5) -- (352.7,165.5) .. controls (355.46,165.5) and (357.7,167.74) .. (357.7,170.5) -- (357.7,210.79) .. controls (357.7,213.55) and (355.46,215.79) .. (352.7,215.79) -- (345.41,215.79) .. controls (342.65,215.79) and (340.41,213.55) .. (340.41,210.79) -- cycle ;
        \draw  [color={rgb, 255:red, 74; green, 144; blue, 226 }  ,draw opacity=1 ][fill={rgb, 255:red, 184; green, 233; blue, 134 }  ,fill opacity=1 ] (342.32,169.36) .. controls (342.32,168.81) and (342.77,168.36) .. (343.32,168.36) -- (354.78,168.36) .. controls (355.34,168.36) and (355.78,168.81) .. (355.78,169.36) -- (355.78,176.53) .. controls (355.78,177.08) and (355.34,177.53) .. (354.78,177.53) -- (343.32,177.53) .. controls (342.77,177.53) and (342.32,177.08) .. (342.32,176.53) -- cycle ;
        \draw  [color={rgb, 255:red, 74; green, 144; blue, 226 }  ,draw opacity=1 ][fill={rgb, 255:red, 184; green, 233; blue, 134 }  ,fill opacity=1 ] (342.32,181.36) .. controls (342.32,180.81) and (342.77,180.36) .. (343.32,180.36) -- (354.78,180.36) .. controls (355.34,180.36) and (355.78,180.81) .. (355.78,181.36) -- (355.78,188.53) .. controls (355.78,189.08) and (355.34,189.53) .. (354.78,189.53) -- (343.32,189.53) .. controls (342.77,189.53) and (342.32,189.08) .. (342.32,188.53) -- cycle ;
        \draw  [color={rgb, 255:red, 74; green, 144; blue, 226 }  ,draw opacity=1 ][fill={rgb, 255:red, 184; green, 233; blue, 134 }  ,fill opacity=1 ] (342.32,193.2) .. controls (342.32,192.64) and (342.77,192.2) .. (343.32,192.2) -- (354.78,192.2) .. controls (355.34,192.2) and (355.78,192.64) .. (355.78,193.2) -- (355.78,200.36) .. controls (355.78,200.92) and (355.34,201.36) .. (354.78,201.36) -- (343.32,201.36) .. controls (342.77,201.36) and (342.32,200.92) .. (342.32,200.36) -- cycle ;
        \draw  [color={rgb, 255:red, 74; green, 144; blue, 226 }  ,draw opacity=1 ][fill={rgb, 255:red, 184; green, 233; blue, 134 }  ,fill opacity=1 ] (342.32,205.03) .. controls (342.32,204.48) and (342.77,204.03) .. (343.32,204.03) -- (354.78,204.03) .. controls (355.34,204.03) and (355.78,204.48) .. (355.78,205.03) -- (355.78,212.2) .. controls (355.78,212.75) and (355.34,213.2) .. (354.78,213.2) -- (343.32,213.2) .. controls (342.77,213.2) and (342.32,212.75) .. (342.32,212.2) -- cycle ;
        
        \draw  [fill={rgb, 255:red, 80; green, 227; blue, 194 }  ,fill opacity=1 ] (252.76,95.5) -- (312.81,122.25) -- (312.81,178.88) -- (252.76,205.63) -- cycle ;
        
        \draw  [fill={rgb, 255:red, 80; green, 227; blue, 194 }  ,fill opacity=1 ] (550.06,205.93) -- (490.01,179.18) -- (490.01,122.55) -- (550.06,95.8) -- cycle ;
        \draw  [fill={rgb, 255:red, 139; green, 87; blue, 42 }  ,fill opacity=1 ] (8.72,26.13) -- (100.35,26.13) -- (100.35,112.85) -- (8.72,112.85) -- cycle ;
        
        \draw  [fill={rgb, 255:red, 139; green, 87; blue, 42 }  ,fill opacity=1 ] (8.39,188.35) -- (99.89,188.35) -- (99.89,280.85) -- (8.39,280.85) -- cycle ;
        \draw    (101.5,70.52) -- (117.5,70.57) -- (129.15,70.61) ;
        \draw [shift={(131.15,70.62)}, rotate = 180.19] [color={rgb, 255:red, 0; green, 0; blue, 0 }  ][line width=0.75]    (10.93,-3.29) .. controls (6.95,-1.4) and (3.31,-0.3) .. (0,0) .. controls (3.31,0.3) and (6.95,1.4) .. (10.93,3.29)   ;
        \draw  [fill={rgb, 255:red, 80; green, 227; blue, 194 }  ,fill opacity=1 ] (132.76,15.17) -- (192.81,41.92) -- (192.81,98.54) -- (132.76,125.29) -- cycle ;
        
        \draw  [fill={rgb, 255:red, 80; green, 227; blue, 194 }  ,fill opacity=1 ] (132.7,180.5) -- (192.75,207.25) -- (192.75,263.88) -- (132.7,290.63) -- cycle ;
        
        \draw    (101.1,234.32) -- (117.1,234.37) -- (128.75,234.41) ;
        \draw [shift={(130.75,234.42)}, rotate = 180.19] [color={rgb, 255:red, 0; green, 0; blue, 0 }  ][line width=0.75]    (10.93,-3.29) .. controls (6.95,-1.4) and (3.31,-0.3) .. (0,0) .. controls (3.31,0.3) and (6.95,1.4) .. (10.93,3.29)   ;
        \draw    (222.9,149.32) -- (238.9,149.37) -- (250.55,149.41) ;
        \draw [shift={(252.55,149.42)}, rotate = 180.19] [color={rgb, 255:red, 0; green, 0; blue, 0 }  ][line width=0.75]    (10.93,-3.29) .. controls (6.95,-1.4) and (3.31,-0.3) .. (0,0) .. controls (3.31,0.3) and (6.95,1.4) .. (10.93,3.29)   ;
        \draw    (192.68,70.52) -- (207.4,70.57) -- (222.33,70.62) ;
        \draw    (193.1,234.32) -- (207.83,234.37) -- (222.75,234.42) ;
        \draw    (222.33,70.62) -- (222.75,234.42) ;
        \draw  [fill={rgb, 255:red, 139; green, 87; blue, 42 }  ,fill opacity=1 ] (8.39,352.02) -- (99.89,352.02) -- (99.89,444.52) -- (8.39,444.52) -- cycle ;
        \draw  [fill={rgb, 255:red, 80; green, 227; blue, 194 }  ,fill opacity=1 ] (132.7,344.17) -- (192.75,370.92) -- (192.75,427.54) -- (132.7,454.29) -- cycle ;
        
        \draw    (101.1,397.98) -- (117.1,398.04) -- (128.75,398.08) ;
        \draw [shift={(130.75,398.08)}, rotate = 180.19] [color={rgb, 255:red, 0; green, 0; blue, 0 }  ][line width=0.75]    (10.93,-3.29) .. controls (6.95,-1.4) and (3.31,-0.3) .. (0,0) .. controls (3.31,0.3) and (6.95,1.4) .. (10.93,3.29)   ;
        \draw  [dash pattern={on 3.75pt off 3.75pt on 3.75pt off 3.75pt}]  (193.53,398.11) -- (208.26,398.16) -- (415.42,398.08) ;
        \draw  [dash pattern={on 3.75pt off 3.75pt on 3.75pt off 3.75pt}]  (415.17,185.83) -- (415.42,398.08) ;
        \draw [shift={(415.17,183.83)}, rotate = 89.93] [color={rgb, 255:red, 0; green, 0; blue, 0 }  ][line width=0.75]    (10.93,-4.9) .. controls (6.95,-2.3) and (3.31,-0.67) .. (0,0) .. controls (3.31,0.67) and (6.95,2.3) .. (10.93,4.9)   ;
        \draw    (459.9,149.32) -- (475.9,149.37) -- (487.55,149.41) ;
        \draw [shift={(489.55,149.42)}, rotate = 180.19] [color={rgb, 255:red, 0; green, 0; blue, 0 }  ][line width=0.75]    (10.93,-3.29) .. controls (6.95,-1.4) and (3.31,-0.3) .. (0,0) .. controls (3.31,0.3) and (6.95,1.4) .. (10.93,3.29)   ;
        \path  [shading=_5xd7kubg7,_yh88qqgdj] (581.39,106.96) -- (673.03,106.96) -- (673.03,193.69) -- (581.39,193.69) -- cycle ; 
         \draw   (581.39,106.96) -- (673.03,106.96) -- (673.03,193.69) -- (581.39,193.69) -- cycle ; 
        
        \draw    (550.9,150.32) -- (566.9,150.37) -- (578.55,150.41) ;
        \draw [shift={(580.55,150.42)}, rotate = 180.19] [color={rgb, 255:red, 0; green, 0; blue, 0 }  ][line width=0.75]    (10.93,-3.29) .. controls (6.95,-1.4) and (3.31,-0.3) .. (0,0) .. controls (3.31,0.3) and (6.95,1.4) .. (10.93,3.29)   ;
        
        \draw (258.32,180) node [anchor=north west][inner sep=0.75pt]  [color={rgb, 255:red, 0; green, 0; blue, 0 }  ,opacity=1 ,rotate=-270] [align=left] {\begin{minipage}[lt]{43.53pt}\setlength\topsep{0pt}
        \begin{center}
        \textbf{VAE}\\\textbf{Encoder}
        \end{center}
        
        \end{minipage}};
        \draw (29.92,226.1) node [anchor=north west][inner sep=0.75pt]   [align=left] {\textcolor[rgb]{1,1,1}{\textbf{Search}}};
        \draw (138.26,268) node [anchor=north west][inner sep=0.75pt]  [color={rgb, 255:red, 255; green, 255; blue, 255 }  ,opacity=1 ,rotate=-270] [align=left] {\begin{minipage}[lt]{47.5pt}\setlength\topsep{0pt}
        \begin{center}
        \textbf{Feature}\\\textbf{Extractor}
        \end{center}
        
        \end{minipage}};
        \draw (138.32,102.67) node [anchor=north west][inner sep=0.75pt]  [color={rgb, 255:red, 255; green, 255; blue, 255 }  ,opacity=1 ,rotate=-270] [align=left] {\begin{minipage}[lt]{47.5pt}\setlength\topsep{0pt}
        \begin{center}
        \textbf{Feature}\\\textbf{Extractor}
        \end{center}
        
        \end{minipage}};
        \draw (21.83,61.69) node [anchor=north west][inner sep=0.75pt]   [align=left] {\textcolor[rgb]{1,1,1}{\textbf{Template}}};
        \draw (19.92,379.43) node [anchor=north west][inner sep=0.75pt]   [align=left] {\begin{minipage}[lt]{50.31pt}\setlength\topsep{0pt}
        \begin{center}
        \textbf{\textcolor[rgb]{1,1,1}{Condition}}\\\textbf{\textcolor[rgb]{1,1,1}{Map}}
        \end{center}
        
        \end{minipage}};
        \draw (138.26,431.67) node [anchor=north west][inner sep=0.75pt]  [color={rgb, 255:red, 255; green, 255; blue, 255 }  ,opacity=1 ,rotate=-270] [align=left] {\begin{minipage}[lt]{47.5pt}\setlength\topsep{0pt}
        \begin{center}
        \textbf{Feature}\\\textbf{Extractor}
        \end{center}
        
        \end{minipage}};
        \draw (343.25,75.75) node [anchor=north west][inner sep=0.75pt]   [align=left] {$\displaystyle \mu $};
        \draw (344.32,148.03) node [anchor=north west][inner sep=0.75pt]   [align=left] {$\displaystyle \sigma $};
        \draw (375.5,112) node [anchor=north west][inner sep=0.75pt]   [align=left] {$\displaystyle \mu +\sigma N[+C]$};
        \draw (544.5,121.43) node [anchor=north west][inner sep=0.75pt]  [color={rgb, 255:red, 0; green, 0; blue, 0 }  ,opacity=1 ,rotate=-90] [align=left] {\begin{minipage}[lt]{43.54pt}\setlength\topsep{0pt}
        \begin{center}
        \textbf{VAE}\\\textbf{Decoder}
        \end{center}
        
        \end{minipage}};
        \draw (590,131) node [anchor=north west][inner sep=0.75pt]   [align=left] {\begin{minipage}[lt]{54.3pt}\setlength\topsep{0pt}
        \begin{center}
        \textbf{\textcolor[rgb]{1,1,1}{Generated}}\\\textbf{\textcolor[rgb]{1,1,1}{Score Map}}
        \end{center}
        
        \end{minipage}};

        \end{tikzpicture}
    }
    \caption{A general diagram of A VAE in single object tracking. The dotted path is the condition used to direct the CVAE, which is only mandatory in the case of a CVAE. In the diagram, $N=\mathcal{N}(0,1)$ is a Gaussian noise vector.}
    \label{fig:vae_diagram}
\end{figure}

A Variational AutoEncoder (VAE) \cite{kingma2013auto-encoding} is a special type of an autoencoder in which the model is prompted to reconstruct a given input $x$ using a learned latent variable $z$, while the latent variable is restricted to have a normal distribution of certain parameters $\mu$ and $\sigma$. The rationale behind such a training paradigm is that it is often easier to manipulate the latent variable. In a typical setting, the learned manifold in the latent space generates new examples that diversify the training dataset. VAE maximizes the data marginal distribution by maximizing its variational lower bound. The lower bound for a data point $x^{i}$ is given by:
\begin{equation}
    \mathcal{L}(\theta, \phi; \boldsymbol{x}^{i}) = \int q_{\phi}(\boldsymbol{z}|\boldsymbol{x})(\log p_{\theta}(\boldsymbol{x}^{i}|\boldsymbol{z}) + \log p_{\theta}(\boldsymbol{x}) - \log q_{\phi}(\boldsymbol{z}|\boldsymbol{x})) \,d\boldsymbol{z}
\end{equation}
where $q_{\phi}(\boldsymbol{z}|\boldsymbol{x})$ refers to the encoder parameterized by $\phi$, $p_{\theta}(\boldsymbol{x}^{i}|\boldsymbol{z})$ refers to the decoder parameterized by $\theta$.
The previous equation yields the training objective for VAE (see \cite{kingma2013auto-encoding}):
\begin{equation}
\min ||\boldsymbol{x} - f_{\phi}(\boldsymbol{z})||_{2} + KL(q_{\phi}(\boldsymbol{z}|\boldsymbol{x})||p_{\theta}(\boldsymbol{z}))
\end{equation}
where $||\boldsymbol{x} - f_{\phi}(\boldsymbol{z})||_{2}$ is the reconstruction error and $KL$ is the Kullback–Leibler divergence \cite{kullback1951information} between the distribution of the encoded data $q_{\phi}(\boldsymbol{z}|\boldsymbol{x})$ and the marginal distribution of the latent variable $p_{\theta}(\boldsymbol{z})$ which is usually assumed to be Gaussian with mean $\mu$ and variance $\sigma$. 
For a long time now, VAEs have been used in a wide variety of computer vision applications ranging from generation \cite{razavi2019generating, cai2019multi, bao2017cvae} to classification \cite{chen2020evolving, chamain2022end} and single object tracking is no exception. A representative diagram of VAE adaptation in single object tracking is shown in Fig.~\ref{fig:vae_diagram}.


\textbf{[SINT++]} Robust Visual Tracking via Adversarial Positive Instance Generation: This work \cite{wang2018sint++}, builds upon Siamese INsTance Search (SINT) \cite{Tao2016Siamese} which uses a Siamese network to search for the first frame template appearance in the next frames of a given video. Overall, this method aims to enhance the robustness of SINT by generating hard positive samples. For that sake, the network comprises two novel components: A Positive Sample Generation Network (PSGN) and a Hard Positive Transformation Network (HPTN). To be succinct, PSGN, the only part implementing a VAE, is discussed here. PSGN is a VAE trained using the VOT dataset \cite{Kristan2018The} to extract a compact object representation from the dataset. This compact manifold $z$ generates more diverse training examples to account for unseen scenarios. After generating diverse samples, these samples are fed into HPTN to be transformed into harder samples, and the final training samples are fed into SINT for training. 

\textbf{[VAE-MCMC]} Robust Visual Tracking Based on Variational Autoencoding Markov Chain Monte Carlo: This research \cite{Kwon2020Robust} proposes tracking with the aid of supporters. Supporters are assumed to be objects whose motion is linked to the target object. They can help locate the target because they are geometrically correlated with it over time. The objective of this method is to find the best target state $X$ and supporters' states $S$ to maximize the marginal probability. The MCMC is used to formulate the joint marginal $p(X, S)$ using proposal and acceptance \cite{Khan2005MCMC-based}. VAE-MCMC decomposes the proposal required for $p(X, S)$ into two proposal terms, each of which is only a function of either the supporters' states or the target state. When dealing with the acceptance phase, the marginal probability of supporters' states $p(S)$ appears in the equation. Although it is possible to reuse MCMC, it is impractical due to the large number of samples required to describe $p(S)$. The authors suggest using variants of VAE that optimize priors \cite{tomczak2018VAE} to marginalize $p(S)$ by decomposing the problem to the optimization of $p_{VAE}(z|S)$, done via the encoder, and $q_{VAE}(S|z)$, done via the decoder.

\subsection{Learning Paradigms}
\label{sec:Learning}

\subsubsection{Self-Supervised Learning}
\label{sec:Self-supervised}

Because deep learning models require a vast amount of labelled training data and acquiring these labelled data is equally exhaustive for humans and machines, self-supervised learning was introduced. Self-supervised learning aims to exploit underlying implicit relations in the data. This paradigm has been explored in video object segmentation \cite{Zhou2023A}, which is usually treated as a twin field to visual tracking with different datasets and applications. Many self-supervised learning methods have been extended to videos based on exploiting temporal relations between video frames. For instance, VFS \cite{Xu2021Rethinking} is self-supervised pre-trained to achieve a video frame correspondence objective on the video dataset Kinetics \cite{Kay2017The}. Moreover, in \cite{Vondrick2018Tracking}, a model is prompted to predict colours in frames based on previously seen video frames. In this survey, we explore self-supervised techniques for visual object tracking. Usually, training with a self-supervised objective entails a cycle-consistency constraint, either explicitly or implicitly. A cycle is defined as a forward-tracking path followed by a backward one. Given a target patch $P_{t}$, a predicted patch by forward tracking $P_{pf}$ and a predicted patch by backward tracking $P_{pb}$, a cycle-consistency objective encourages that:
\begin{equation}
    \label{eq:cycle-consistency}
    P_{pf} \equiv P_{t}
\end{equation}
where the identical shape and location of the equivalent patches characterize the equivalence.


\textbf{[Joint-Task]} Joint-task Self-supervised Learning for Temporal Correspondence: The main assumption in this method \cite{Li2019Joint-task} is that for any two consecutive-in-time feature maps, there has to be a linear transformation represented by an affinity matrix that maps the features and locations from one feature map to the other. This method obtains the affinity matrix by a simple dot product followed by a softmax operation on the corresponding feature maps. Learning to match features requires sampling the corresponding frames, which involves many issues, such as staticity and the absence of the target frame. To alleviate the effect of these issues, a two-stage framework is introduced, where two affinity matrices $A_{fp}$ and $A_{pp}$ are learned. These affinity matrices represent the linear transformation applied to the shape and the location. $A_{fp}$ represents a transformation mapping the reference frame to the target patch, while $A_{pp}$ represents a transformation mapping the target and current patches. First, a region-level localization network uses two greyscale inputs: reference patch and search frame. This phase aims to determine the location transformation that, when applied using the learned $A_{fp}$, a new patch of the target object is located in the search frame. Second, fine-grained matching is applied using the colour information. During this phase, $A_{pp}$ is obtained through the target and cropped search patches. Furthermore, an autoencoder is employed to generate the coloured target patch with the encoder prompted by the current search patch and the decoder mixing the latent representation with the $A_{pp}$ affinity matrix. The training objective encompasses two regularization conditions. First, the cycle-consistency constraint, which in the context of this method can be achieved by the orthogonality of the affinity matrices alongside the location matching from the predicted and original locations of the object in the same frame cyclically. Second, the concentration constraint necessitates pixels from the same object not to drift from its center, moving as one unit over time. 

\textbf{[CycleSiam]} Self-supervised Object Tracking with Cycle-consistent Siamese Networks: This method \cite{Yuan2020Self-supervised} employs the cycle-consistency described by equation \ref{eq:cycle-consistency} in SiamRPN \cite{li2018high}. The basic idea is to establish forward and backward tracking passes for many frames at various timesteps. The loss is calculated upon the completion of each cycle. The loss function contains two components, one for bounding box consistency and another for tracked object mask consistency. It is worth mentioning that this method is trained using labels (i.e., supervised learning) to identify the object location in the initial frame. However, the training was done on a general-purpose ImageNet challenge \cite{Olga2015ImageNet}, and its performance was exclusively evaluated on the Visual Object Tracking (VOT) dataset \cite{Kristan2018The}.

\textbf{[UniTrack]} Do Different Tracking Tasks Require Different Appearance Models?: Although this method \cite{Wang2021Do} does not train an end-to-end network in a self-supervised tracking task, it paved the way for generic self-supervised pre-trained models to unify different tracking tasks. This work suggests that all tracking tasks such as SOT \cite{Wu2013Online}, VOS \cite{Perazzi2016A}, MOT \cite{Milan2016MOT16}, MOTS \cite{Voigtlaender2019MOTS}, and PoseTrack \cite{Andriluka2018PoseTrack} require a base model that transforms images into informative feature maps that capture discriminative patterns. The base models, which have been empirically proven effective, are self-supervised models trained on general-purpose image and video datasets. Experiments conducted by the authors prove that VFS \cite{Xu2021Rethinking} is the best performant base model. Moving on with the base model embeddings, the tracking tasks are categorized into propagation (SOT, VOS) and association tasks (MOT, MOTS, PoseTrack). Cross-correlation is adopted as the relation modelling technique for our scope of propagation tasks. Different heads are employed for the final output layer to meet the requirements of different tasks (e.g., bounding box head for SOT, mask head for VOS, etc.). 

\textbf{[Crop-Transform-Paste]} Self-Supervised Tracking via Target-Aware Data Synthesis: As the title suggests, this work \cite{Li2022Self-Supervised} provides tracking models with different data augmentations that are suited for the visual tracking problem. Given a training sequence, first, the target is cropped out of the initial frame using the ground truth bounding box. Following that, the target undergoes transformations that simulate the challenges present in the single object tracking datasets. These challenges include motion blur, occlusion, scale change, illumination variation, and deformation. The challenges can be represented by transformations, including blur, cutout, rescale, colour jitter, and shear, respectively. Lastly, the cropped and transformed target is pasted on a random location of random images of the same sequence. Henceforth, more data examples are obtained with information miming the different conditions existing in single object tracking datasets. 

\textbf{[ETC]} Exploiting Temporal Coherence for Self-supervised Visual Tracking by Using Vision Transformer: The main hypothesis in this work \cite{Zhu2022Exploiting} is that there exist characteristics in the target object that are invariant to viewpoint change and deformation. The objective of this method is to exploit the intrinsic supervisory signals in videos. The tracking problem is formulated assuming that two types of Samples exist: Unlabelled Samples (ULS) and Few Labelled Samples (FLS). On the one hand, while training using ULS, no ground truth location of the target object is given. On the other hand, the initial location of the target object is given in FLS. Although frames are sampled equally for ULS to exploit the potential of unlabelled images better, they are sampled variably for FLS to use the label information. Two types of relations between frames are assumed to exploit the temporal coherence: intra-frame (frame-level) and inter-frame (sequence-level). Training sequences are represented by palindromes of frames where some $T$ frames are reflected and appended to the training sequence to contain $\{F_{i}, F_{i+1}, F_{i+2}, ..., F_{T}, ..., F_{i+2}, F_{i+1}, F_{i}\}$ to train a cycle-consistent transformer. The main contribution in this work, the cycle-consistent transformer, implements equation \ref{eq:cycle-consistency} using an encoder-decoder architecture. An essential step before the transformer is encoding all frames in the palindrome using a CNN. After that, the encoder encodes the template representation of $F_{i}$ and the search representation of $F_{i+1}$ and feeds a joint representation for the decoder, which predicts a representation for the template in $F_{i+1}$ which in turn is used as a template representation in the next transformer block. Furthermore, the training objective is two-fold. One is to ensure the final result cycle consistency, and the other is to ensure the consistency of intermediate results. For instance, in a given palindrome, where template ground truths are $\{F^{gt}_{i}, F^{gt}_{i+1}, F^{gt}_{i+2}, ..., F^{gt}_{T}, ..., F^{gt}_{i+2}, F^{gt}_{i+1}, F^{gt}_{i}\}$ and predicted targets in search images are $\{F^{s}_{i}, F^{s}_{i+1}, F^{s}_{i+2}, ..., F^{s}_{T}, ..., F^{gt}_{i+2}, F^{s}_{i+1}, F^{s}_{i}\}$, the model is trained such that:
\begin{equation}
\label{eq:intermediate-cycle-consistency}
    F^{s}_{i} \approx F^{gt}_{i}, ..., F^{s}_{T} \approx F^{gt}_{T}
\end{equation}

\textbf{[QCT]} Consistency-based Self-supervised Visual Tracking by Using Query Communication Transformer: The same method of \cite{Zhu2022Exploiting} is reintroduced in this work \cite{Zhu2023Consistency-based} changing the transformer architecture to propagate the template and search embeddings as query information between instances of a query interaction module. This method leverages the transformer architecture to enhance the model's performance based on \cite{Zhu2022Exploiting}. This enhancement is realized by the transformer's capability to capture long-term dependencies represented by the query token. 

\subsubsection{Unsupervised Learning}
\label{sec:Unsupervised}

In this category, the labels given in the tracking dataset are entirely excluded from the training process. As opposed to self-supervised learning, supervisory signals are not obtained from the ground truth in any training phase. Instead, the gradient signals come from an intrinsic source of information to the data. For instance, given an image in a tracking setting, an unsupervised method aims to teach the model how to track an arbitrary object or pattern, in some cases, not only a given labelled object. 

\textbf{[UDT]} Unsupervised Deep Tracking: This work \cite{Wang2019Unsupervised} trains an unsupervised Siamese model using the cycle-consistency concept introduced in section \ref{sec:Self-supervised}. The notable difference between self-supervised methods and this method is that they use the labels to localize the object initially. In contrast, this method uses random patterns in the images to align with the unsupervised learning paradigm. To further improve the localization accuracy, more frames can be inserted in the cycle where the result template from a previous frame is used to search in the next frame. Moreover, a cost-sensitive loss is introduced to eliminate noisy and uninformative patches that may result from random cropping. To account for noisy patches, all patches are ranked for their loss values, and the highest $10\%$ of them do not contribute to the final loss. Considering informative patches, a weight is given according to motion by calculating the mean squared distance between the ground truth synthesized labels and the response map. If the distance is larger than a certain threshold, then the patch is probably informative. 

\textbf{[PUL]} Progressive Unsupervised Learning for Visual Object Tracking: This method \cite{Wu2021Progressive} trains Siamese tracking networks like SiamFC \cite{bertinetto2016fully-convolutional} and SiamDW \cite{zhang2019deeper} without supervision. It consists of three main stages: Background discrimination, temporal mining, and temporal correspondence. Background discrimination uses EdgeBox \cite{Zitnick2014Edge} to extract object proposals, which are then augmented inspired by SimCLR \cite{Chen2020A} and MoCo \cite{he2020momentum}, and trained to distinguish between objects using contrastive learning. Hard negative mining is employed to enhance feature learning. Temporal mining tracks patterns across frames, and temporal correspondence ensures robustness to noise using a noise-robust loss based on target map likelihood assumptions.

\subsection{Meta Learning}
\label{sec:MetaLearning}

By exposing the model to diverse tracking scenarios within a comprehensive dataset, meta-learning empowers it to "learn how to learn." During tracking, the model leverages this meta-knowledge to rapidly adapt to a new target within a limited number of initial frames \cite{Finn2017Model-Agnostic}. This adaptation may involve fine-tuning internal parameters or selecting an optimal tracking strategy based on the object's specific characteristics.

Meta-learning offers several advantages for single object tracking. First, it alleviates the need for extensive training data encompassing every object category. Second, it enables the tracker to handle unforeseen circumstances through on-the-fly adaptation. Finally, meta-learning has the potential to enhance real-time performance by facilitating the acquisition of efficient adaptation strategies. While ongoing research is crucial, meta-learning demonstrates promise as a powerful tool for developing robust and adaptable single-object tracking algorithms.

\textbf{[Meta-Tracker]} Fast and Robust Online Adaptation for Visual Object Trackers: This method \cite{Park2018Meta-Tracker} addresses the problems of initializing the set of trainable parameters $\theta_{0}$ and finding an adaptive learning rate $\alpha$. Concretely, given random initial parameters $\theta_{0}^{0}$ and $\alpha^{0}$, an initial frame $x_{0}$ and a frame interval $\delta$, the model is prompted using the initial frame $x_{0}$. Afterwards, after one iteration, another frame $x_{\delta}$ is sampled as a testing frame to evaluate the quality of the predicted parameters. Subsequently, the model runs $T$ iterations of optimization to find optimal $\theta_{0}^{T}$. These values are considered constant when training the tracker to obtain a new set of parameters $\theta_{i}$ where $i\neq0$. For any subsequent frame, the same algorithm can be used in updating $\alpha$. However, it is found that $\alpha$ tends to be large when optimized for very few frames, thus leading to unstable training. Consequently, $\alpha$ is fixed for the subsequent frames after it is obtained through optimization in the initial step.

\textbf{[DIMP]} Learning Discriminative Model Prediction for Tracking: This research \cite{Bhat2019Learning} depends on a combination of meta-learning approaches. The model comprises a feature extractor, a discriminative convolutional layer and a model predictor. At a glance, the overall idea is that once target and search features are extracted, target features are fed into the model predictor, which infers the weights of the discriminative convolutional layer. Shifting towards meta-learning techniques, first, a meta-representation choice is made for the loss function. The loss function is designed to ensure discrimination between foreground and background. Given a feature map output of a feature extraction network, the loss aims to learn the parameters of the loss function such that the feature map matches a ground truth object center map. Furthermore, the ground truth map is considered part of the meta-representation. Second, the learning rate is learned during the optimization in a data-driven approach using the steepest descent. Third, for initiating the weights of the discriminative convolutional layer, a model-agnostic meta-learning approach \cite{Finn2017Model-Agnostic} is used. The bounding box is predicted using the overlap maximization approach by \cite{Danelljan2019ATOM}. During online tracking, augmented views are extracted from the first annotated frame and used as input to the model predictor to initialize and optimize the convolutional discriminator. The discriminator is optimized for other frames when a distractor peak is detected or after passing a certain number of frames. A memory of the 50 best-predicted frames is maintained at each run to ensure the optimal weight update for the discriminator through the model predictor.

\textbf{[CLNet]} Compact Latent Network for Fast Adjusting Siamese Trackers: This work \cite{dong2020clnet} hypothesizes that the Siamese trackers tend to misclassify samples of the testing (tracking) data due to the bad choice of the decision boundary that separates the training data. The rationale behind the tracker's behaviour is that it does not sample enough decisive positive and negative samples from the available testing data in the first frame. In other words, the Siamese trackers lack the ability to fully leverage the annotations in the first frame, resulting in missing information about the whole sequence. To alleviate this effect, many positive and negative samples must be drawn from the first frame alongside the training samples from the other frames. Considering the easy solution of training on training samples and fine-tuning on testing samples, the small amount of testing data may lead to overfitting, and the same decision boundary problem may appear again in the training data. This work experiments with the previously mentioned assumptions on SiamRPN++ \cite{Li2019SiamRPN++}. The proposed solution to the aforementioned problem is a compact latent network that adapts the weights of the last layer of the classification and regression branches to the testing samples. The compact latent network takes positive and negative samples from a hidden feature map, then predicts their mean $\mu$ and standard deviation $\sigma$ and finally concatenates $\mu^{+}$, $\mu^{-}$, $\sigma^{+}$, $\sigma^{-}$ into a compact latent variable that is used to update the last layer's weights. The primary assumption in this work is that a Gaussian distribution represents the testing and training samples. Therefore, if we can incorporate the information about the distribution while determining the decision boundary, then the decision boundary will consider almost the whole population in the distribution, not only the sampled points. From a meta-learning perspective, this method can be seen as MAML \cite{Finn2017Model-Agnostic} with a carefully designed subnetwork that considers assumptions about the tracking problem.

\textbf{[MGA-Net]} Meta-graph Adaptation for Visual Object Tracking: This method \cite{Wu2021Meta-Graph} adjusts loss gradients of specific layers in a chosen tracker before weight updates. This process involves embedding gradients into a low-dimensional space and combining them with target representations. Gradient embedding constructs a graph where vertices represent gradients and edges represent weight similarities. A graph neural network transforms this graph into a semantic representation concatenated with target embeddings. A filter adaptation network combines gradients and template embeddings, encoding and decoding them to obtain refined gradients. Meta-training involves running the tracker and updating gradients iteratively using embedding and adaptation modules. Online tracking optimizes computational resources by updating gradients for every $T$ frames based on reliable results from previous frames, mitigating overfitting.

\textbf{[RTS]} Robust Visual Tracking by Segmentation: This work \cite{paul2022robust} assumes that visual tracking with a bounding box prior is not robust to variations such as occlusion because the object boundary is ill-defined by the parameters of the bounding box. To serve the purpose of segmentation-centred tracking, the model adapts a segmentation method called Learning What to Learn (LWL) \cite{bhat2020learning}. Mainly, RTS consists of two branches: a segmentation branch based on LWL and an instance localization via the classification branch. Both branches adapt similar architecture but for different tasks. On the one hand, the segmentation branch aims to learn high-quality discriminative masks. On the other hand, the instance localization branch aims to remedy the lack of robustness of the segmentation masks when dealing with long and challenging tracking sequences. In the segmentation branch, a meta-learning model learns the weights of a segmentation feature encoder given a previous memory of successful segmentation masks. In the instance localization branch, another meta-learning model learns the weights of an instance classification model using DIMP \cite{Bhat2019Learning}, which adopted a loss function for robustness in the case of tracking videos. The feature maps produced by both branches are aggregated through a weighted addition and fed into a decoder. The decoder, conditioned by both branches, finally outputs a segmentation map of the tracked object. Both branches are trained jointly, with the segmentation branch updated on every iteration and the instance classification updated on the first iteration only of the inner optimization iterations. 

\subsection{Continual Learning}
\label{sec:ContinualLearning}

Typically, continual learning approaches are intended for long-term tracking, which is outside the scope of this survey. However, the continual learning methodology is useful even for long, short-term sequences. 

\textbf{[TrackingViaContinualLearning]} Long-Term Visual Object Tracking via Continual Learning: This work \cite{Zhang2019Long-Term} applies continual learning via knowledge distillation \cite{hinton2015distilling}. Namely, the model consists of two main components: a dynamic memory and a continual learning module. The dynamic memory is divided into positive samples short-term, positive samples long-term, and negative samples short-term sub-memories. The continual learning module is trained through a classification loss and a continual knowledge distillation loss. In the warm-up phase of training, the model is trained using the positive and negative samples of the short-term memory only and a cross-entropy classification loss is computed. In the main stage of training, the joint training, the previously warmed-up model weights are frozen, and a new model is instantiated to be adapted to the positive samples from the long-term memory. In this stage, the old model is used as a ground-truth provider, and the cross entropy between the predictions of the new model and the old model's predictions is minimized as a knowledge distillation component in the total loss. The complete knowledge distillation algorithm is the same as the Learning without Forgetting (LwF) algorithm \cite{li2017learning}. The last piece of the method is updating the long-term memory of positive samples. This is done by comparing the highest scaler value of the response map to a threshold. The update happens if the value is less than a threshold. The memory is also updated every 10 frames. 

\textbf{[Con]} Visual Tracking by Adaptive Continual Meta-Learning: This research \cite{Choi2022Visual} exploits the same paradigm of knowledge distillation. In particular, the model comprises a meta-learner and a tracking baseline. The meta-learner aims to learn the following hyperparameters: the learning rate, the weighting parameters for knowledge distillation, current instance components of the loss, and one current-instance-loss-specific hyperparameter. For that purpose, the given dataset is divided into four episodes containing consecutive frames. The first episode is responsible for the initial adaptation of model parameters. Using the remaining episodes, the model runs inner optimization iterations on a given episode to obtain an initial prediction of the model weights while distilling the knowledge of the previous episode. Formally, given episodes $\mathcal{D}_{1}$, $\mathcal{D}_{2}$, $\mathcal{D}_{3}$ and $\mathcal{D}_{4}$, the inner optimization output is the model parameters $\theta_{1}$, $\theta_{2}$ and $\theta_{3}$ where the focal loss \cite{Lin2017Focal} is employed to learn $\theta_{1}$ and the loss employed to learn $\theta_{2}$ and $\theta_{3}$ is a linear combination of the knowledge distillation and the focal losses. For the outer optimization, the model performs meta-training using all weights obtained from the inner optimization steps. Furthermore, the outer loop aims to find not only the optimal parameters for the model itself but also to optimize the meta-learner network weights. In this stage, different combinations of the preset episodes are used to enhance generalization. Also, the loss used in this stage is the focal loss with fixed hyperparameters in contrast to the inner loop focal loss.

\subsection{Domain Adaptation}
\label{sec:DomainAdaptation}
In single object tracking, domain adaptation involves training a model on a source dataset encompassing various tracking tasks under varying conditions. This comprehensive training fosters the extraction of domain-agnostic features, such as motion patterns and object shapes that are crucial for tracking across different scenarios. During tracking in a new, unseen target domain (e.g., a video with distinct lighting or background), the model can then adapt its internal parameters or select the optimal tracking strategy based on the specific characteristics of the target object and its environment. This adaptation process leverages the generalizable knowledge acquired from the source domain while accounting for the target domain's unique statistical properties.

\textbf{[MDNet]} Learning Multi-Domain Convolutional Neural Networks for Visual Tracking: This method \cite{Nam2016Learning} is the first method to tackle the domain adaptation aspect of visual object tracking. The model encompasses two main components: A CNN feature extractor and a sequence-dependent fully connected layer. The feature extractor is trained offline to gain domain-independent knowledge, while the last fully connected layer is trained online to attain domain knowledge. Using region proposals of a detection algorithm, the model classifies them as positive or negative targets, thus defining an easier problem for the bounding box head to solve. Moreover, the model incorporates hard negative mining \cite{Sung1998Example-based} to learn non-trivial discrimination between the target and the background. Finally, the bounding box is obtained via a bounding box regression head using features from a convolutional layer following \cite{girshick2014rich}.

\textbf{[CODA]} Real-Time Deep Tracking via Corrective Domain Adaptation: This method \cite{Li2019Real-Time} deals with the domain adaptation requirement by presenting a detection-aided tracking methodology. The model consists of three main components: A tracker, a single-shot detector and an algorithm for tracking output adjustment via the detector. First, the tracker is based on hierarchical correlation filter \cite{Ma2015Hierarchical} where outputs from different hierarchy convolutional layers are inputs to kernelized correlation filter \cite{Henriques2015High-Speed} mini trackers. These mini trackers are prompted to output a coarse tracking result. The main contribution of the tracker is the additional domain adaptive layers, which are multi-resolution convolutional layers applied to the feature maps before being fed to the KCF trackers. These domain adaptive layers learn the receptive field required for each task through learning importance weights for concatenated outputs of different convolutional filters $3\times3$, $5\times5$, and $7\times7$ following the design principal introduced in Inception CNN \cite{szegedy2015going}. After that, bounding box heads are applied to the output of the KCF trackers and an intermediate bounding box is drawn. Finally, the model adjusts the tracker output by modifying the dimensions of the output bounding box. The bounding box area is adjusted based on the average of the areas of positive bounding boxes obtained from the single-shot detector. A detector bounding box must have IoU and confidence scores that pass predefined thresholds to be considered positive. 

\section{Experimental Analysis}
\label{sec:Experimental}

\begin{table*}
\centering
\caption{Summary of the reported performance measures. The datasets involved are: GOT-10k \cite{Huang2021GOT-10K}, LaSOT \cite{Fan2019LaSOT}, LaSOT\textsubscript{ext} \cite{Fan2021LaSOT}, TrackingNet \cite{Muller2018TrackingNet}, OTB50 \cite{Wu2013Online}, OTB100 \cite{ning2016object}, COCO \cite{Lin2014Microsoft}, ILSVRC \cite{Olga2015ImageNet}, ALOV++ \cite{Smeulders2014Visual}, UAV123 \cite{mueller2016benchmark}, NUS-PRO \cite{Li2016NUS-PRO}, NFS \cite{Galoogahi2017Need}, NYUv2 \cite{Silberman2012Indoor}, YouTube-VOS \cite{Xu2018YouTube-VOS}, YouTube-BB \cite{Real2017YouTube-BoundingBoxes}, Kinetics300 \cite{Kay2017The}, and DAVIS \cite{Perazzi2016A}. Please refer to section \ref{sec:datasets} for a brief overview of the reported datasets.}
\label{tab:performance_split}
\resizebox{\textwidth}{!}{
\begin{tabular}{c|c|cccccccccccccc} 
\hline
\multirow{2}{*}{Method}         & \multirow{2}{*}{Training Datasets}                   & \multicolumn{3}{c}{GOT-10k} & \multicolumn{3}{c}{LaSOT} & \multicolumn{3}{c}{LaSOT\textsubscript{ext}} & \multicolumn{3}{c}{TrackingNet} & OTB50 & OTB100  \\ 
\cline{3-16}
                                &                                                      & AO   & $SR_{0.5}$ & $SR_{0.75}$ & AUC  & $P_{Norm}$ & P       & AUC  & $P_{Norm}$ & P                          & AUC  & $P_{Norm}$ & P             & AUC   & AUC     \\ 
\hline
\multicolumn{16}{c}{Autoregressive}                                                                                                                                                                                                                 \\ 
\hline
\multicolumn{2}{c|}{\cite{wei2023autoregressive}}                                      &      &          &           &      &          &         &      &          &                            &      &          &               &       &         \\ 
\hline
ARTrack256                      & \multirow{3}{*}{GOT-10k, LaSOT, TrackingNet}         & 73.5 & 82.2     & 70.9  & 70.4 & 79.5  & 76.6   & 46.4 & 56.5     & 52.3 & 84.2 & 88.7     & 83.5         &   -    &   -      \\        
\cline{1-1}\cline{3-16}
ARTrack384                      &                                                      & 75.5 & 84.3     & 74.3  & 72.6 & 81.7  & 79.1   & 51.9 & 62.0     & 58.5  & 85.1 & 89.1     & 84.8
       &   -    &   -      \\ 
\cline{1-1}\cline{3-16}
ARTrack-L384                    &                                                      & 78.5 & 87.4     & 77.8  & 73.1 & 82.2  & 80.3   & 52.8 & 62.9     & 59.7  & 85.6 & 89.6     & 86.0       &   -    &   -      \\
\hline
\end{tabular}
}
\end{table*}

\begin{table*}
\centering
\caption{Summary of the reported performance measures on the VOT challenge dataset from 2016 to 2020 \cite{Kristan2016The, Kristan2017The, Kristan2018The, Kristan2019The, Kristan2020The}.}
\label{tab:performance_vot}
\resizebox{\textwidth}{!}{
\begin{tabular}{c|c|ccccc} 
\hline
\multirow{2}{*}{Method}                & \multirow{2}{*}{Training Datasets}                 & VOT16 & VOT17 & VOT18 & VOT19 & VOT20  \\ 
\cline{3-7}
                                       &                                                    & EAO   & EAO   & EAO   & EAO   & EAO    \\ 
\hline
\multicolumn{7}{c}{Memory-based sequence models}                                                                                     \\ 
\hline
\cite{yang2017recurrent}               & ILSVRC                                             & -     & 22.2  & -     & -     & -      \\ 
\hline
\cite{gao2019graph}                    & ILSVRC                                             & -     & 27.4  & -     & -     & -      \\ 
\hline
\cite{zhao2021cooperative}             & GOT-10k                                            & 37.5  & -     & 39.8  & -     & -      \\ 
\hline
\cite{zhang2023attention}              & Online                                             & 36.8  & -     & 25.8  & -     & -      \\ 
\hline
\cite{wang2023robust}                  & GOT-10k, LaSOT, TrackingNet, COCO                  & -     & -     & -     & -     & 50.1   \\
\hline
\end{tabular}
}
\end{table*}

\begin{table*}[ht]
\caption{A brief overview of the selected datasets used in reporting the different performance metrics of the discussed trackers. Other mentioned training datasets in Tab.~\ref{tab:performance_split} are briefly discussed in section \ref{sec:datasets}.}
\label{tab:datasets}
\resizebox{\textwidth}{!}{%
%
                      \\ \bottomrule
\end{tabular}%
}
\end{table*}

In this section, we report the performance of the discussed trackers on different single object tracking datasets. We report selected tracking performance measures and the datasets used for training each of the discussed models in Tab.~\ref{tab:performance_split}. We also report the Expected Average Overlap metric used in the renowned Visual Object Tracking (VOT) challenge in Tab.~\ref{tab:performance_vot}.  

\subsection{Selected Datasets}
\label{sec:datasets}

A set of the most popular single object tracking datasets was selected to report the test performance of the surveyed trackers. The test sequences include challenging situations such as Object Color Change (OCC), Fast Motion (FM), In-Plane Rotation (IPR), Aspect Ratio Change (ARC), Illumination Variation (IV), OCClusion (OCC), Similar OBject (SOB), Low-Resolution Objects (LRO), Viewpoint Change (VC), Full OCclusion (FOC), Out-of-View (OV), Fast Camera Motion (FCM), DEFormation (DEF), Low Resolution (LR), Absolute Motion (AM), Motion Blur (MB), Out-of-Plane rotation (OPR), target ROTation (ROT), Background Clutter (BC), Partial OCclusion (POC), abrupt Camera Motion (CM), Scene COmplexity (SCO), Scale Variation (SV), MOtion Change (MOC). A detailed definition of each attribute can be found in \cite{Muller2018TrackingNet, Fan2021LaSOT}. A brief comparison of the selected testing datasets is presented in Tab.~\ref{tab:datasets}. We also report the datasets each model was trained on without specifically being single-object-tracking datasets. \textbf{COCO} \cite{Lin2014Microsoft} is a large scale object detection and segmenetation dataset with $200k$ images. \textbf{ILSVRC} \cite{Olga2015ImageNet} is the ImageNet Large Scale Visual Recognition Challenge that uses the ImageNet dataset, the largest open-access dataset with $10M$ images intended for classification. It is also extended to other tasks, e.g. video object detection. \textbf{ALOV++} \cite{Smeulders2014Visual} contains $315$ sequences from the Amsterdam Library of Ordinary Videos focused on robustness evaluation with one tracking challenge per sequence. It provides annotations for every $5$ frame per sequence. \textbf{UAV123} \cite{mueller2016benchmark} is a single-object Unmanned-Aerial-Vehicle-perspective video dataset with $123$ sequences. It also includes a subset for long-term tracking. \textbf{NUS-PRO} \cite{Li2016NUS-PRO} was proposed as a new tracking challenge using a dataset of $365$ YouTube sequences. \textbf{Kinetics} \cite{Kay2017The} is a large scale human action recognition with approximately $500k$ sequences. \textbf{NFS} \cite{Galoogahi2017Need} is the highest frame rate ($240$ FPS) tracking dataset with $100$ sequences and $380k$ frames. \textbf{NYUv2} \cite{Silberman2012Indoor} is an indoor segmentation and depth estimation dataset with $1449$ labelled examples with aligned RGB and depth images. \textbf{YouTube-VOS} \cite{Xu2018YouTube-VOS} is a large-scale video object segementation dataset with arround $5k$ sequences and $190k$ segementation annotations. \textbf{YouTube-BB} \cite{Real2017YouTube-BoundingBoxes} is a general purpose large-scale dataset comprises $380k$ sequences with $5.6M$ bounding box annotations. \textbf{DAVIS} \cite{Perazzi2016A} is a densely annotated (per-frame annotations) video object segmentation dataset. It contains $50$ sequences with mask annotations per object instance. 

Due to the significance of the single object tracking problem, An ongoing annual challenge is held, namely the VOT challenge \cite{Kristan2016The, Kristan2017The, Kristan2018The, Kristan2019The, Kristan2020The}. The VOT challenge utilizes the VOT dataset, which consists of $60$ sequences that involve many challenging scenarios in tracking. The challenge does not require a specific training set and does not provide one. However, the challenge prohibits the usage of $1000$ sequences from GOT-10k \cite{Huang2021GOT-10K} because they were used in refreshing the sequences of the VOT20 dataset. It is worth mentioning that the optimal VOT bounding box is the one that surrounds the object while containing minimal background pixels, which is in contrast to other datasets that require a box with its sides aligned with the image sides. The main judging metric employed by the challenge is the Expected Average Overlap discussed in Section \ref{sec:metrics}. We 

\subsection{Performance Metrics}
\label{sec:metrics}

\subsubsection{Average Overlap (AO)} The AO is the main adopted metric by GOT-10k \cite{Huang2021GOT-10K}. It measures the mean average overlap between the ground truth bounding boxes and the predicted bounding boxes. Instead of directly averaging the overlaps in all sequences, a per-class average is calculated, and each class average overlap is weighted according to the number of class samples. Finally, all averages are reduced to a per-dataset overlap. Calculating the AO this way accounts for the class imbalance in the dataset. 

\subsubsection{Success Rates ($SR_{0.5}$ and $SR_{0.75}$)} The secodary measure adopted by GOT-10k \cite{Huang2021GOT-10K} to indicate accuracy-robustness trade-off. The success rate is the percentage of predicted bounding boxes that exceed an overlap threshold. The threshold can be either $0.5$ or $0.75$. As for the AO, the success rates are calculated per class and then reduced per dataset to accommodate the imbalance. Uniquely, the success rate calculation is performed three times and averaged to get a robust measure. 

\subsubsection{Area Under the Success Curve (AUC)} This is the main performance measure adopted by LaSOT \cite{Fan2019LaSOT, Fan2021LaSOT}, TrackingNet \cite{Muller2018TrackingNet} and OTB \cite{Wu2013Online, ning2016object}. First, the success curve is obtained by defining a $0 \sim 1$ range of thresholds as the $x$-axis and another $0 \sim 1$ range of success ratios. A success ratio at a certain threshold is the number of frames in the predicted bounding boxes that overlap with the ground truth bounding boxes greater than the threshold. The area under the curve is reported after forming the success curve for different thresholds.

\subsubsection{Precision (P)} Tracker precision is the distance between the centres of the predicted and ground truth bounding boxes in pixels. The reported precision, which is reported for LaSOT \cite{Fan2019LaSOT, Fan2021LaSOT} and TrackingNet \cite{Muller2018TrackingNet}, is the ratio of the frames that has a precision greater than a certain threshold (usually $=20$ pixels). 

\subsubsection{Normalized Precision ($P_{Norm}$)} This measure was introduced in TrackingNet \cite{Muller2018TrackingNet} and adopted in LaSOT \cite{Fan2019LaSOT, Fan2021LaSOT}. It mainly aims to cancel the effect of the object and image dimensions in calculating the precision. This is done by dividing the precision per frame by the ground truth bounding box size. 

\subsubsection{Expected Average Overlap (EAO)} Adopted by VOT challenge datasets since 2015 \cite{Kristan2016The, Kristan2017The, Kristan2018The, Kristan2019The, Kristan2020The} and calculated by averaging the average overlaps on a very large set of sequences. This accommodates the high variance of the AO.  

\subsection{Cumulative Remarks} Although performance comparison between different trackers is a key to understanding them, we put forth that reporting the performance for the different trackers is for completeness. We do not mean to compare different trackers that emerged in distant years, having different backbones and leveraging different computation capabilities. Rather, we report the performance to observe the evolution in single object tracking methodologies. Please refer to Section~\ref{sec:Taxonomy} for more information about how different algorithms might affect the performance. 

\section{Discussion and Future Directions}
\label{sec:Discussion}
\begin{figure}    
    \scalebox{0.5}{
    \includegraphics[width=2\columnwidth]{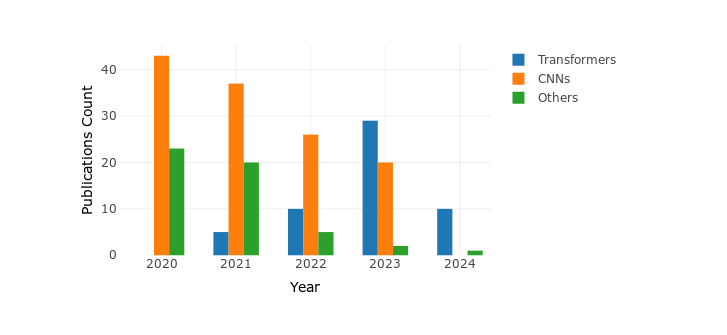}
    }
    \caption{A breakdown of the number of publications in single object tracking that utilize transformers, CNNs, or any other approach such as DCF.}
    \label{fig:TransVSCNN}
\end{figure}
This survey provides a comprehensive outlook on emerging theoretical techniques in deep learning and their application in single-object tracking. In this section, we summarize the main takeaways and suggest future pathways for single-object tracking based on the trends we observe throughout the survey. 

Although this survey focuses on the new applications of non-traditional theoretical techniques in deep learning, it is worth mentioning that a big portion of the research in object tracking is shifting toward the \textbf{vision transformer paradigm}. To this end, we investigate the increase in the employment of vision transformers in single object tracking from $2020$ to $2024$ in Fig.~\ref{fig:TransVSCNN}. In this investigation, six conferences, CVPR, ICCV, ECCV, AAAI, NeurIPS, and WACV, and four journals, TPAMI, TIP, IJCV, and TCSVT, among the top-tier venues in computer vision, are adopted. We count the research publications on single object tracking in these $10$ venues. We refer to the publications that utilize a transformer, either purely or jointly with a CNN, as "Transformers." For pure CNN-based approaches we refer to them as "CNNs." We refer to DCF-based or any method other than transformer-based or CNN-based methods as "Others." From Fig~\ref{fig:TransVSCNN}, it can be deduced that although the CNN-based approaches dominated single object tracking research before $2023$, these approaches are losing the research interest to the transformer-based approaches. A similar trend can be seen with methods other than CNN-based. Overall, the current research in single object tracking is centred around transformer-based approaches. Many factors can justify this trend. One factor is that the transformer architecture is inherently of a bigger receptive field than that of CNNs, thus adapting to changes in the scale of the target object.  Another factor is the introduction of large-scale single object tracking datasets such as GOT-10k \cite{Huang2021GOT-10K} and LaSOT \cite{Fan2019LaSOT}, which suit the transformer architecture. A third factor is the explicit attention paid to the template and search tokens. Moreover, by design, the transformers do not have the strong inductive bias that the CNNs have. These factors enable a more generic representation of the relation between the template and the search images. The vision transformer's powerful generic representation capability is particularly useful for the single object tracking task that requires capturing the critical relations from only one annotated example at tracking time. Moreover, the attention mechanism effectively handles long-range dependencies, crucial for single-object tracking, especially long-term tracking. 

Another trend we observe is the \textbf{incorporation of natural language}. With the current state of the Large Language Models, it is natural to have related applications in single object tracking. Mainly, the usage of LLMs-based techniques falls into two categories; one is natural language tracking \cite{Li2017Tracking, Li2022Cross-Modal, Zhou2023Joint} in which a tracker is prompted to track an object given a specification provided in natural language. The other category, which is fairly recent, is the usage of the natural language in enhancing the tracking model performance \cite{Li2023CiteTracker}. The latter category is interesting because it enforces a tracker to associate a generic language description with a specific visual appearance, leveraging information that is usually unemployed. These methods pave the way for generic cross-modal models that integrate multiple tasks. 

As per this work, we observe the importance of \textbf{explicit learning of temporal correspondence}. The current state-of-the-art techniques sequence models effectively learn the temporal cues by autoregression. Following the success of autoregressive Large Language Models (LLMs) \cite{yang2019generalized, Brown2020Language}, it is natural to extend the concept of autoregression in single object tracking to calculate appearance models for a dynamically changing template object. This way turns out to be effective when applied to a one-stream tracking framework such as \cite{ye2022joint}.

Furthermore, we highlight the significance of the \textbf{backbone pretraining} in a supervised or self-supervised manner. More recent techniques employ pretraining in a supervised way such as \cite{Zhao2023Representation, Wu2023DropMAE} where a task of joint reconstruction of template and search images helps the model understand the relationship between the template and its surroundings as well as the temporal variations in a target object implicitly. In another context of self-supervised unlabelled pretraining, existing works emphasize the importance of the cycle consistency constraint as a non-trivial pretext task that serves visual tracking. 

Another observation is the powerful concept of \textbf{generative models}. In particular, the GAN-based methods enhance the tracking model discrimination abilities. However, generative models such as diffusion models have still not been discovered. Despite landing in object detection \cite{Chen2023DiffusionDet} and multi-object tracking \cite{luo2023diffusiontrack}, its application in single-object tracking is still far behind. As for the time being, diffusion is only used to generate more samples but not for an end-to-end single object tracking framework.

As future pathways, we see that more research will employ learning autoregressive temporal cues alongside learning robust visual representations. This is due to the increasing amount of research that learns either of them separately. A joint approach can be implemented by penalizing a reconstruction error alongside maximizing the likelihood of the next autoregressive bounding box token. We also anticipate that a stronger pretext task for self-supervised tracking will be adopted. This is especially possible with the introduction of many novel pretext tasks for learning discriminative visual features and robust temporal features. An example of a novel pretext task is Mixed Autoencoder \cite{Chen2023Mixed}, where the autoencoder is prompted to minimize similar object's embeddings and maximize the different object embeddings in the same image. Another novel pretext task is introduced in No More Shortcuts (NMS) \cite{dave2023shortcuts} where a model is prompted to predict the position of a frame in a given sequence via two concurrent tasks, one is the skip rate and the other is whether the frame is out of order. This kind of temporal dependency capturing in a pretraining phase enables a holistic understanding of the changes that may occur to an object in a tracking setting. 

Finally, a promising direction is harnessing \textbf{diffusion models} to object tracking by following the method proposed by \cite{Chen2023DiffusionDet} with the diffusion denoising process done on the bounding boxes.

\backmatter








\section*{Declarations}

\begin{itemize}
\item \textbf{Funding.} The authors did not receive support from any organization for the submitted work.
\item Competing interests: The authors have no relevant financial or non-financial interests to disclose.
\item \textbf{Ethics approval and consent to participate.} Not applicable
\item \textbf{Consent for publication.} The authors consent to publish this work in the International Journal of Machine Learning and Cybernetics.
\item \textbf{Data availability.} Not applicable.
\item \textbf{Materials availability.} Not applicable.
\item \textbf{Code availability.} Not applicable.
\item \textbf{Author contribution.} Conceptualization: Omar Abdelaziz, Mohamed Shahata, Mohamed Mohamed; Methodology: Omar Abdelaziz; Formal analysis and investigation: Omar Abdelaziz; Writing - original draft preparation: Omar Abdelaziz; Writing - review and editing: Omar Abdelaziz, Mohamed Shahata, Mohamed Mohamed; Supervision: Mohamed Shehata
\end{itemize}

\end{document}